\documentclass[11pt]{article}

\usepackage[preprint]{acl}

\usepackage{times}
\usepackage{latexsym}
\usepackage{amsmath,amssymb}
\usepackage{enumitem}
\usepackage{multirow}
\usepackage{minitoc}
\usepackage{titletoc}
\usepackage{booktabs}
\usepackage{colortbl}
\usepackage{pifont}

\usepackage{case_layout}

\definecolor{lightpurplerow}{RGB}{242,236,250}
\usepackage{tabularx}
\usepackage{tikz}

\usepackage[T1]{fontenc}

\usepackage[utf8]{inputenc}

\usepackage{microtype}

\usepackage{inconsolata}
\usepackage{algorithm}
\usepackage{algpseudocode}

\usepackage{graphicx}
\usepackage{hyperref}
\definecolor{Heartvermillion}{RGB}{159,54,58}
\hypersetup{
    colorlinks=true,
    linkcolor=Heartvermillion,
    citecolor=Heartvermillion,
    filecolor=magenta,      
    urlcolor=Heartvermillion,
    }

\title{Of Errors and Echoes: Responsibility-Aware Credit Assignment for Tool-Use Agents}

\author{
 \textbf{Xingjian Diao$^{1*}$},
 \textbf{Tianyu Yang$^{2*}$},
 \textbf{Wenjun Huang$^3$},
 \textbf{Xiangchi Yuan$^4$},
 \textbf{Chunhui Zhang$^1$}
 \\
 \textbf{Xingchen Zhao$^5$}\textbf{,}
 \textbf{Weiyi Wu$^1$}\textbf{,}
 \textbf{Haotian Xu$^6$}\textbf{,} 
 \textbf{Soroush Vosoughi$^1$}\textbf{,}
 \\
 \textbf{Xiangliang Zhang$^{2\dagger}$}\textbf{,}
 \textbf{Jiang Gui$^{1\dagger}$}
 \\
 $^1$Dartmouth College \quad
 $^2$University of Notre Dame \quad
 $^3$University of California, Irvine \\
 $^4$Georgia Institute of Technology \quad
 $^5$Northeastern University \quad
 $^6$Independent Researcher
 \\
   \texttt{xingjian.diao.gr@dartmouth.edu},
     \texttt{tyang4@nd.edu}
}

\begin{document}
\maketitle

\newcommand\blfootnote[1]{%
  \begingroup
  \renewcommand\thefootnote{}\footnote{#1}%
  \addtocounter{footnote}{-1}%
  \endgroup
}

\blfootnote{$^*$Equal contribution, listed in random order.}
\blfootnote{$^\dagger$Corresponding authors.}

\begin{abstract}

Reinforcement learning improves tool-use agents, yet fine-grained credit assignment remains challenging even with step-level feedback. An observed error may be inherited from an earlier decision, while multiple responsible decisions may share the same downstream consequence or participate in later recovery. We introduce Error-Chain-Aware Optimization (ECHO), a responsibility-aware credit assignment framework that separates responsibility identification from credit allocation. A frozen judge combines retrospective trajectory analysis with decision-time assessment from the policy-visible prefix to distinguish newly introduced errors, inherited consequences, missed corrections, and successful corrections. ECHO then counts shared downstream consequences once across responsible decisions and constrains rewards for repairing self-induced errors through source-funded recovery, producing bounded action-level supervision alongside terminal supervision. Experiments on six diverse tool-use benchmarks at both 4B and 8B scales show strong and consistent improvements over the compared baselines, with ECHO achieving the highest score on five of six benchmarks at each scale. These results support decision responsibility as a useful basis for fine-grained credit assignment in tool-use agents. Our code is available at \url{https://github.com/xid32/ECHO}.
\end{abstract}

\begin{figure}[t]
    \centering
    \includegraphics[width=\columnwidth]{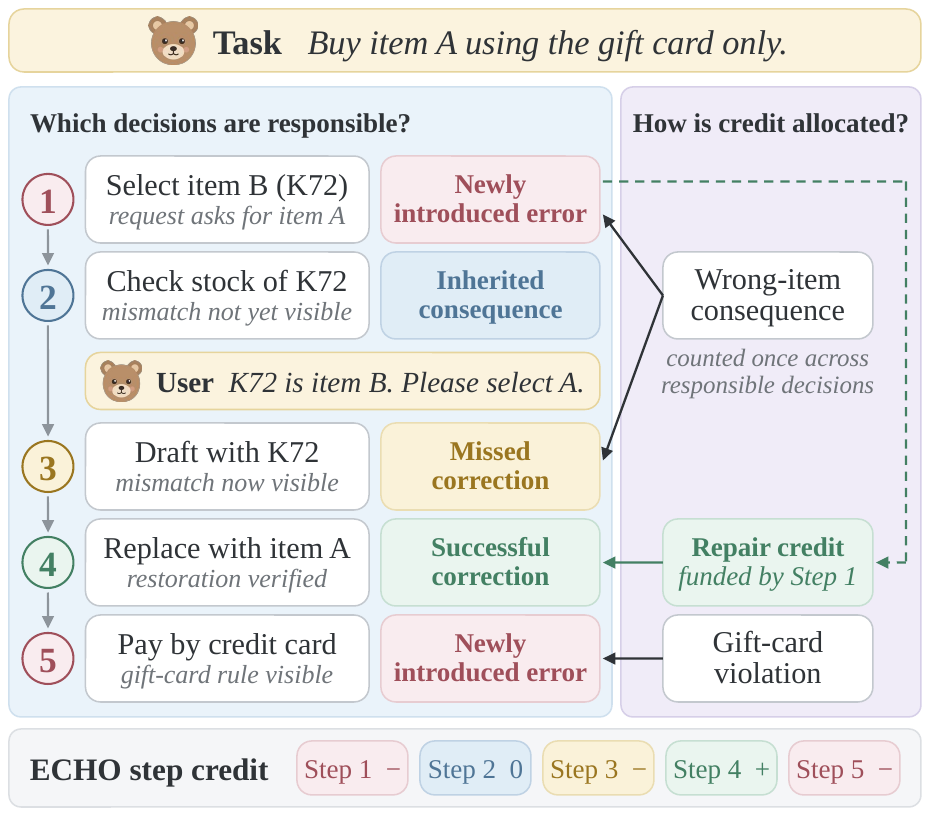}
    \par\vspace{-0.3cm}
    \caption{
    ECHO on an error chain.
    \emph{Left:} Responsibility depends on decision-time visibility: Step~2 inherits the error before the mismatch is visible, while Step~3 misses a correction after disclosure.
    \emph{Right:} Credit allocation counts the wrong-item consequence once across Steps~1 and~3 and funds Step~4's repair credit from Step~1.
    }
    \vspace{-0.6cm}
    \label{fig:echo-teaser}
\end{figure}
\section{Introduction}
\label{sec:introduction}

Tool-use agents retrieve information and modify environments through sequential decisions
\citep{yao2023react,schick2023toolformer,qin2024toolllm},
where each tool call shapes the state and constraints that later calls depend on
\citep{zhou2024webarena,yao2025taubench,lu2025toolsandbox}.
Reinforcement learning can improve such agents \citep{wang2025ragen,qian2025toolrl}, but terminal rewards indicate only whether a trajectory succeeds, not which decisions led to the outcome.
Step-level or process-level feedback localizes where errors occur
\citep{lightman2024verify,wang2024mathshepherd,zhang2025lessons,zheng2025processbench},
yet the step at which an error surfaces need not be the decision responsible for it: an error may be inherited from an earlier decision, multiple responsible decisions may share a downstream consequence, and later decisions may repair the error.
These dependencies raise two questions: \textit{(i)} which decisions are responsible, and \textit{(ii)} how credit should be allocated among them.

Figure~\ref{fig:echo-teaser} illustrates both questions. 
The user asks for item A and requires payment by gift card only, but the agent selects item B with identifier \texttt{K72} (Step~1), checks the stock of \texttt{K72} (Step~2), keeps using it after the mismatch is disclosed (Step~3), restores item A only later (Step~4), and finally pays by credit card (Step~5). 
Steps~2 and~3 show that responsibility depends on decision-time visibility: both act on the wrong item, but only Step~3 acts after the mismatch is visible, so Step~2 is an inherited consequence whereas Step~3 is a missed correction.
Steps~1,~3, and~4 show why credit must be allocated across the error chain rather than independently to each step.
Steps~1 and~3 contribute to the same wrong-item consequence, so charging each the full penalty would count it twice.
Step~4 deserves repair credit, but if that credit is decoupled from the source error, the policy may profit from creating and then fixing its own mistake.
Step~5, by contrast, is a separate error with its own consequence.

We formalize these cases as four responsibility events.
A newly introduced error (\textsc{New}) creates a deviation from a task requirement that visible information should have prevented.
An inherited consequence (\textsc{Inh}) carries forward an earlier deviation before the discrepancy becomes visible.
A missed correction (\textsc{Miss}) leaves a visible, correctable deviation unaddressed.
A successful correction (\textsc{Corr}) verifiably restores a violated requirement or required precondition.
In Figure~\ref{fig:echo-teaser}, Steps~1--4 illustrate these events in order, while Step~5 introduces another \textsc{New} error. 
Beyond this example, we map diagnostics from our six evaluation benchmarks to the evidence each event requires (\S\ref{app:taxonomy}) and show all four events in untrained-policy trajectories (\S\ref{app:base-cases}).

To address both questions, we introduce \textbf{ECHO} (\textbf{E}rror-\textbf{CH}ain-Aware \textbf{O}ptimization), which assigns credit across error chains rather than to independent steps.
For responsibility identification, a frozen judge uses the full trajectory only to propose candidate events and verify error propagation and successful repair, while each selected action is judged from its policy-visible prefix, preventing hindsight leakage.
For credit allocation, shared-consequence accounting charges each consequence once across its responsible decisions, and source-funded recovery limits repair credit to a budget derived from the source error's penalty.
The resulting bounded action-level corrections are optimized alongside the terminal reward.
Our contributions are:

\begin{itemize}[leftmargin=*]
    \item We identify that the step where an error surfaces need not be the decision responsible for it, and formalize this with four responsibility events: newly introduced errors, inherited consequences, missed corrections, and successful corrections.
    \item We propose ECHO, which assesses responsibility at decision time and allocates credit over error chains, addressing hindsight leakage, double counting of shared consequences, and over-crediting self-repair.
    \item At both 4B and 8B, ECHO leads the compared methods on five of six tool-use benchmarks, with case studies illustrating behavioral improvements and ablations supporting each component.
\end{itemize}
\section{Related Work}
\label{sec:related-work}

\noindent
\textbf{Tool-Use Agent Learning.}
Reinforcement learning has substantially improved tool-use agents through reward design, multi-turn rollouts, and simulated users \citep{qian2025toolrl,feng2026retool,zhao2025muarl,dong2025arpo,zhou2024archer,wen2024poad,luo2025lightning}, as well as through scaled training environments \citep{song2026envscaler,wang2026awm,wang2025klearagentforge,li2025simia}. These methods primarily optimize trajectory-level outcomes or scale the environments used for training. ECHO is complementary, addressing how supervision is distributed across individual agent decisions when errors propagate, become observable, or are later corrected.

\noindent
\textbf{Process Supervision and LLM Critics.}
Step-level feedback provides denser supervision than terminal rewards \citep{lightman2024verify,wang2024mathshepherd,zhang2025lessons,zheng2025processbench}, and reward models and LLM critics extend this idea to search and tool-use agents \citep{xia2025agentrm,li2026toolrm,xi2026agentprm,tan2025tarl,tan2026hcapo,zhou2025sweetrl,zhang2026criticsearch}. Such signals localize problematic steps, but localization alone does not establish decision responsibility: an observed error may be inherited from an earlier decision, and judging a decision with later evidence can penalize it for information unavailable when it was made. ECHO therefore uses the full trajectory only to propose candidate events and verify propagation or restoration, assessing each decision independently from its policy-visible prefix.

\noindent
\textbf{Credit Assignment and Error Attribution.}
Beyond identifying responsible decisions, a complementary question is how delayed outcomes should be allocated among them. Reward redistribution and recent agent methods estimate step or turn contributions from delayed returns, trajectory structure, counterfactual rollouts, or turn-level value estimates \citep{arjona2019rudder,meulemans2023cocoa,zhang2023grd,feng2025gigpo,li2026salt,wang2026cwgrpo,gandhi2026draco,meng2026craft,tao2026trace,liao2026tcpo,zhan2026slcagrpo}, while failure-attribution studies identify root-cause errors for post-hoc diagnosis \citep{zhang2025whowhen,zhu2025agentdebug}. Conceptually related to structural accounts of responsibility \citep{chockler2004responsibility}, ECHO explicitly separates responsibility identification from credit allocation: shared-consequence accounting counts each consequence once across its responsible decisions, and source-funded recovery bounds self-induced repair credit by the budget of its originating error.

\begin{figure*}[t]
    \centering
    \includegraphics[width=\textwidth]{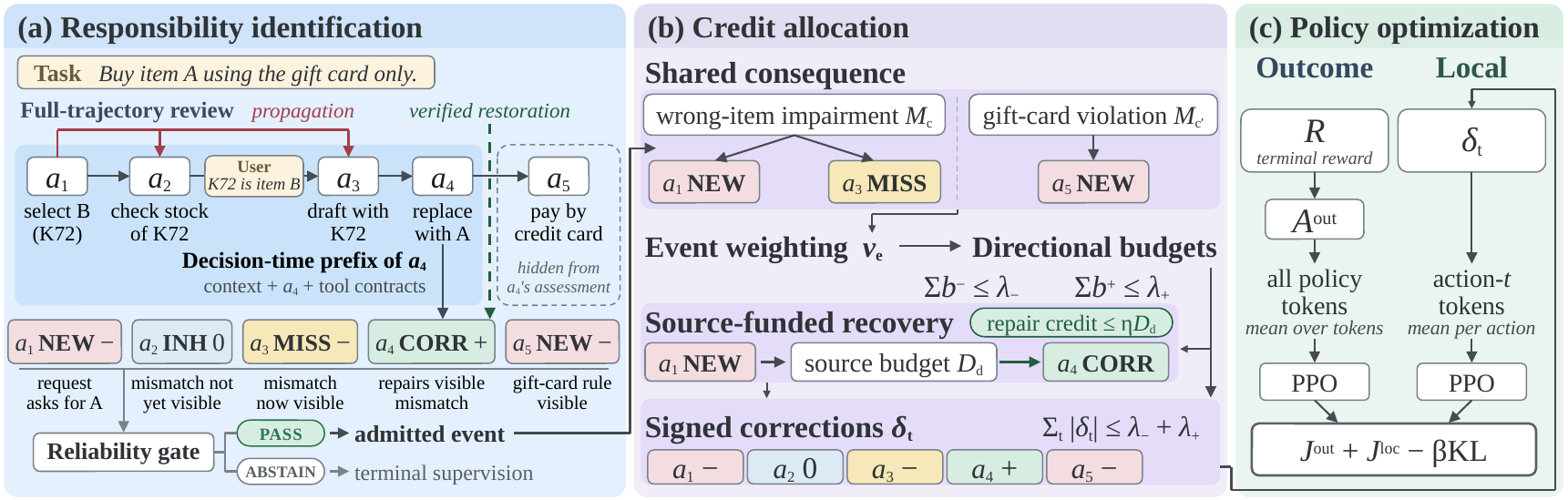}
    \par\vspace{-6pt}
    \caption{
    Overview of ECHO.
    (a) Full-trajectory review proposes candidate events, propagation links, and restoration evidence, while decision-time assessment classifies each selected action as \textsc{New}, \textsc{Inh}, \textsc{Miss}, or \textsc{Corr}.
    (b) Shared consequences are counted once across responsible actions, then weighted, normalized, and adjusted by source-funded recovery to yield signed action-level corrections $\delta_t$.
    (c) Terminal and local supervision are optimized through separately clipped outcome-level and action-level branches with KL regularization.
    }
    \vspace{-0.3cm}
    \label{fig:echo-methods}
\end{figure*}

\section{ECHO}
\label{sec:methods}

ECHO complements the terminal reward with action-level corrections $\delta_t$ (Figure~\ref{fig:echo-methods}, Algorithm~\ref{alg:echo}).
A frozen judge classifies selected actions as \textsc{New}, \textsc{Inh}, \textsc{Miss}, or \textsc{Corr} using decision-time context (\S\ref{sec:assessment}); ECHO then counts each shared penalty once, caps self-induced repair credit by its source penalty (\S\ref{sec:accounting}--\ref{sec:budgets}), and optimizes $\delta_t$ alongside the terminal advantage (\S\ref{sec:optimization}).
In Figure~\ref{fig:echo-teaser}, Steps~1 and~3 split the wrong-item penalty, Step~2 receives no local correction because it is an inherited consequence (\textsc{Inh}), Step~4's repair credit is funded by Step~1's penalty, and Step~5 is charged separately.

\begin{algorithm}[ht]
\small
\caption{One ECHO update}
\label{alg:echo}
\begin{algorithmic}[1]
\Require policy $\pi_\theta$, frozen judge $\mathcal J$, batch of tasks
\State Sample $G$ trajectories per task; record $R_i$ and $h_{i,t}$
\For{each trajectory $i$}
    \State Propose events, links, and restorations from the full trajectory
    \For{each selected action $t$ ($\leq K$)}
        \State Assess type $z$ from decision-time context only
    \EndFor
    \State Retain events passing structural checks and $g_e$
    \State Allocate and weight events \hfill Eqs.~\eqref{eq:ownership}--\eqref{eq:weight}
    \State Normalize into budgets $b_e^s$ \hfill Eq.~\eqref{eq:budget}
    \State Cap self-induced repair credit \hfill Eqs.~\eqref{eq:principal}--\eqref{eq:escrow}
    \State Aggregate corrections $\delta_{i,t}$ \hfill Eq.~\eqref{eq:delta}
\EndFor
\State Compute $A_i^{\mathrm{out}}$ and optimize $J_{\mathrm{ECHO}}$ \hfill Eqs.~\eqref{eq:outcome}--\eqref{eq:objective}
\end{algorithmic}
\end{algorithm}

\subsection{Problem Setup and Responsibility Events}
\label{sec:events}

Trajectory $i$ of policy $\pi_\theta$ on task $x$ consists of actions $a_{i,t}$, observations $o_{i,t}$, and terminal reward $R_i$.
The decision-time context $h_{i,t}$ records the information visible to the policy before $a_{i,t}$, after any context truncation or memory retrieval.
We represent checkable task requirements as predicates $f\in\mathcal F_x$, such as the requested item and the gift-card rule in Figure~\ref{fig:echo-teaser}. 
The judge outputs responsibility events
\begin{equation}
 e=\langle t,f,d,z\rangle,
\label{eq:event}
\end{equation}
where $t$ is the assessed action, $f$ the affected predicate, $d$ the deviation origin (self-induced, authenticated external, or unknown), and $z$ the responsibility type.
Each event also stores supporting evidence pointers and, for a missed correction, a feasible alternative response (\S\ref{app:audit}).
A single action may produce multiple events, for example by repairing one predicate while violating another; their action-level coefficients are aggregated in Eq.~\eqref{eq:delta}.

\subsection{Responsibility Identification from Decision-Time Prefixes}
\label{sec:assessment}

\paragraph{Responsibility types.}
Local responsibility is assessed from evidence visible in $h_{i,t}$.
\ding{182}~A newly introduced error (\textsc{New}) is a distinct decision defect evident from a fact or constraint visible to the policy.
\ding{183}~A missed correction (\textsc{Miss}) requires an existing deviation, visible evidence of the discrepancy, and a feasible permitted response that the action fails to take.
\ding{184}~An inherited consequence (\textsc{Inh}) reflects an earlier deviation with explicit provenance but involves neither a new defect nor a missed correction.
\ding{185}~A successful correction (\textsc{Corr}) is a legitimate response to a visible deviation, with execution evidence confirming restoration of the predicate or a required precondition. Unverified repair claims receive no positive credit.
Actions without events retain only terminal supervision, and for external failures ECHO assesses only the agent's response to the incident.
Duplicate events are merged, while ambiguous or conflicting labels resolve to a single supported type or abstention.

\paragraph{Independent prefix assessment.}
The judge plays two roles.
Given the full trajectory, it proposes candidate events, propagation links, and restoration evidence.
For at most $K$ selected actions, it then independently assesses responsibility from $(h_{i,t},a_{i,t})$ and the tool contracts visible at that time, without access to the proposals, hidden state, or future observations.
Under fixed decoding, the local assessment of a selected action is therefore invariant to the trajectory's suffix.
Hindsight informs candidate selection, propagation links, restoration evidence, and event weighting, but not the decision-time responsibility judgment. 
We set $K=8$, with up to six proposed actions and two non-proposed sentinels that probe for missed events (\S\ref{app:audit}).

\paragraph{Selective supervision.}
Each event must pass structural checks on its evidence pointers, visibility, temporal order, provenance, and nonempty policy span.
A binary gate $g_e$ further admits only evidence strata whose semantic error rate, estimated from independent annotations, falls below a calibrated bound (\S\ref{app:calibration}).
Events that fail either check receive only terminal supervision.

\subsection{Shared-Consequence Accounting}
\label{sec:accounting}

\paragraph{Constraint-impairment proxy.}
Before inspecting outcomes, a fixed task adapter extracts nonredundant requirement predicates $\mathcal F_x$ from public instructions and tool contracts, with masses $\omega_f$ that sum to one and are uniform by default.
Predicate versions and primitive subfacts share the mass of their underlying requirement; obligations outside this inventory contribute only through the direct-defect term below.
Repeated manifestations of the same predicate's impairment form one \emph{consequence} $c$ with mass
\begin{equation}
 M_c=\omega_{f(c)}s_c,\qquad \sum_c M_c\leq1,
\label{eq:mass}
\end{equation}
where severity $s_c$ is $0$ for unconfirmed, $1/2$ for reversible, and $1$ for terminal or irreversible impairment.
Recovery does not erase the source defect or reduce its recorded maximum severity.

\paragraph{Divisible ownership.}
Let $P_c$ contain the responsibility slots linked to consequence $c$, including uncertified slots.
After reserving a fraction $\zeta_c$ for uncertain provenance, each slot receives an equal share:
\begin{equation}
 a_{ec}=\begin{cases}
 \dfrac{1-\zeta_c}{|P_c|},& e\in P_c,\ |P_c|>0,\\
 0,&\text{otherwise},
 \end{cases}
\label{eq:ownership}
\end{equation}
where $\zeta_c$ is $0$ for complete recorded provenance, $1/2$ for unresolved additional sources, and $1$ for unsupported or conflicting ownership.
The reserved fraction and the shares of rejected events are not reassigned. 
Because one trajectory cannot identify each decision's causal contribution (\S\ref{app:attribution-limits}), ECHO adopts equal shares as a neutral convention: in Figure~\ref{fig:echo-teaser}, the source error at Step~1 and the missed correction at Step~3 each bear half of the wrong-item consequence, so it is counted only once.

\paragraph{Event weights.}
For a negative event,
\begin{equation}
 \widehat\ell_e^-=
 \kappa\,\iota_e+(1-\kappa)\sum_c M_c a_{ec},
\label{eq:loss}
\end{equation}
where $\iota_e\in\{0,1\}$ indicates a certified distinct decision defect and $0<\kappa\leq1$.
The direct term counts each certified defect once, including defects that are later repaired.
For a successful correction $r$, let first-restoration shares $z_{rc}$ satisfy $\sum_r z_{rc}\leq1$, and define
\begin{equation}
 \widehat\ell_r^+=\kappa+(1-\kappa)\sum_c M_c z_{rc}.
\label{eq:restoration}
\end{equation}
Let $\mathcal F_e\subseteq\mathcal F_x$ contain the predicates reached from event $e$ through validated propagation links, namely explicit reuse of corrupted values or invalidated conditions.
For corrections, $\mathcal F_e$ is restricted to the repaired origin's footprint.
For either event direction,
\begin{equation}
 p_e=\sum_{f\in\mathcal F_e}\omega_f,
 \qquad v_e=g_e\widehat\ell_e[1+\alpha p_e],
\label{eq:weight}
\end{equation}
where $\widehat\ell_e$ denotes the corresponding negative-event or correction base weight and $0\leq\alpha\leq1$.
Since $p_e\leq1$, propagation increases an event's weight by at most a factor of $1+\alpha$.

\subsection{Source-Funded Recovery Budgets}
\label{sec:budgets}

Let $\mathcal E_i^-$ contain the admitted \textsc{New} and \textsc{Miss} events of trajectory $i$, and $\mathcal E_i^+$ its execution-verified \textsc{Corr} events.
\textsc{Inh} and abstained events receive no auxiliary budget and retain only terminal supervision.
Weights are normalized into directional budgets
\begin{equation}
 b_e^s=\lambda_s
 \frac{v_e}{\max\{1,\sum_{u\in\mathcal E_i^s}v_u\}},
 \qquad s\in\{-,+\},
\label{eq:budget}
\end{equation}
so that each direction totals at most $\lambda_s$, while weight sums below one retain their original scale.

\paragraph{Source-funded recovery.}
Without further constraints, a policy could earn positive credit by repairing errors it introduced itself.
ECHO therefore funds such repairs from the penalty assigned to their source.
For a self-induced origin $d$, the source penalty is
\begin{equation}
 D_d=\sum_{e:\,z_e=\textsc{New},\,d(e)=d}b_e^-,
\label{eq:principal}
\end{equation}
with each \textsc{New} charge assigned to exactly one origin.
Let $\mathcal R_d$ contain the eligible restorations of origin $d$, with total provisional credit
$B_d^+=\sum_{r\in\mathcal R_d}b_r^+$.
For $B_d^+>0$, set
$\gamma_d=\min\{1,\eta D_d/B_d^+\}$, and otherwise set $\gamma_d=0$.
The paid correction coefficients are
\begin{equation}
 \widetilde b_r^+=\begin{cases}
 \gamma_d b_r^+,&d\text{ self-induced},\\
 b_r^+,&d\text{ authenticated external},\\
 0,&\text{otherwise},
 \end{cases}
\label{eq:escrow}
\end{equation}
where $0<\eta\leq1$ bounds self-induced repair credit to a fixed fraction of its source penalty.
Partial repairs of the same origin share this budget, unused credit is discarded, and each authenticated external incident permits one recovery entitlement.
In Figure~\ref{fig:echo-teaser}, Step~1's penalty funds the repair at Step~4, while the payment violation at Step~5 is charged separately.

\paragraph{Action-level corrections.}
Aggregating events on each action yields
\begin{equation}
 \delta_{i,t}=
 \sum_{r\in\mathcal E_{i,t}^+}\widetilde b_r^+
 -\sum_{e\in\mathcal E_{i,t}^-}b_e^-,
\label{eq:delta}
\end{equation}
so that
$\sum_t|\delta_{i,t}|\leq\lambda_-+\lambda_+$.
The resulting $\delta_{i,t}$ enters the local objective (\S\ref{sec:optimization}) without further rescaling.
Positive and negative event records are retained even when their coefficients cancel on the same action.

\begin{table*}[t]
\centering
\small
\setlength{\tabcolsep}{3pt}
\renewcommand{\arraystretch}{1.15}
\setlength{\doublerulesep}{1.5pt}
\setlength{\aboverulesep}{0pt}
\setlength{\belowrulesep}{0pt}
\begin{tabularx}{\textwidth}{c|>{\raggedright\arraybackslash}X|cccccc}
\hline\hline
\multicolumn{2}{c|}{} & \multicolumn{6}{c}{\textbf{Overall score}} \\[1pt]
\cmidrule(l){3-8}
\multicolumn{2}{c|}{\textbf{Model}} & \rule{0pt}{2.6ex}\textbf{BFCL-v3} & \textbf{BFCL-v4} & \textbf{$\tau$-bench} & \textbf{$\tau^2$-Bench} & \textbf{ToolSandbox} & \textbf{ACEBench} \\
\hline
\multirow{4}{*}{4B} & Qwen3~\citep{yang2025qwen3} & 25.38 & 38.61 & 33.44 & 15.83 & 52.89 & 55.28 \\
 & EnvScaler~\citep{song2026envscaler} & 37.25 & \underline{52.92} & \underline{48.78} & \underline{33.05} & 62.25 & \textbf{63.33} \\
 & Arctic-AWM~\citep{wang2026awm} & \underline{38.89} & 49.55 & 26.58 & 22.57 & \underline{68.24} & 60.58 \\
\rowcolor{lightpurplerow}
\cellcolor{white} & \textbf{ECHO (ours)} & \textbf{41.24} & \textbf{53.71} & \textbf{50.77} & \textbf{35.81} & \textbf{76.36} & \underline{62.46} \\
\hline
\multirow{7}{*}{8B} & Qwen3~\citep{yang2025qwen3} & 28.88 & 41.58 & 36.65 & 26.44 & 59.74 & 60.00 \\
 & EnvScaler~\citep{song2026envscaler} & 41.38 & 54.20 & 44.22 & 35.10 & \underline{68.85} & 58.33 \\
 & Arctic-AWM~\citep{wang2026awm} & \underline{43.70} & 52.50 & 37.64 & 33.45 & 60.44 & 54.93 \\
 & Klear-AgentForge~\citep{wang2025klearagentforge} & 37.84 & 47.88 & \underline{49.10} & 35.91 & 66.06 & \textbf{60.70} \\
 & MUA-RL~\citep{zhao2025muarl} & 14.63 & 26.84 & 43.00 & 30.20 & 50.00 & 53.30 \\
 & Simia-RL~\citep{li2025simia} & 41.64 & \underline{55.72} & 44.27 & \underline{40.16} & 62.25 & 58.33 \\
\rowcolor{lightpurplerow}
\cellcolor{white} & \textbf{ECHO (ours)} & \textbf{49.17} & \textbf{56.91} & \textbf{50.59} & \textbf{40.56} & \textbf{74.42} & \underline{60.08} \\
\hline\hline
\end{tabularx}
\vspace{-0.2cm}
\caption{
Overall results on six tool-use benchmarks at the 4B and 8B scales, spanning multi-turn function calling, conversational task completion, and stateful tool execution. 
For BFCL-v3, we report the overall score on the Multi-Turn (MT) subset. For ACEBench, we report results on the Agent subset.
Scores are percentages. \textbf{Bold} and \underline{underlined} values mark the best and second-best results within each scale.
}
\vspace{-0.4cm}
\label{tab:main-results}
\end{table*}

\subsection{Length-Balanced Policy Optimization}
\label{sec:optimization}

Let $q_{i,k}$ indicate a loss-eligible policy token and $m_{i,t,k}$ its membership in action $t$, with $m_{i,t,k}\leq q_{i,k}$ and $\sum_t m_{i,t,k}\leq1$ so that local action spans do not overlap.
Set $L_i=\sum_k q_{i,k}$ and $B_{i,t}=\sum_k m_{i,t,k}$.
Actions with $B_{i,t}=0$ contribute no local loss, while trajectories with $L_i=0$ contribute no policy loss.
For trajectories sampled from the same task, the group-relative outcome advantage is \citep{shao2024deepseekmath}
\begin{equation}
 A_i^{\mathrm{out}}=\frac{R_i-\mu_x}{\sigma_x+\varepsilon},
\label{eq:outcome}
\end{equation}
where $\mu_x$ and $\sigma_x$ are the mean and population standard deviation of the group rewards and $\varepsilon>0$ is a small constant.
Equal-reward groups therefore have zero outcome advantage but may still receive local corrections.

Let $\rho_{i,k}$ be the current-to-behavior token likelihood ratio, and define the clipped surrogate \citep{schulman2017ppo}
\begin{equation}
 \phi_\epsilon(\rho,A)=
 \min\{\rho A,\operatorname{clip}(\rho,1-\epsilon,1+\epsilon)A\}.
\label{eq:phi}
\end{equation}
ECHO maximizes
\begin{align}
 J_i^{\mathrm{out}}&=\frac1{L_i}\sum_kq_{i,k}
 \phi_\epsilon(\rho_{i,k},A_i^{\mathrm{out}}),\label{eq:jout}\\
 J_i^{\mathrm{loc}}&=\sum_{t:B_{i,t}>0}\frac1{B_{i,t}}
 \sum_km_{i,t,k}\phi_\epsilon(\rho_{i,k},\delta_{i,t}),\label{eq:jloc}\\
 J_{\mathrm{ECHO}}&=\frac1N\sum_i
 [J_i^{\mathrm{out}}+J_i^{\mathrm{loc}}-\beta K_i],\label{eq:objective}
\end{align}
where $K_i$ is the trajectory-mean reference-policy KL penalty and $N$ counts contributing trajectories.
Judge outputs, rewards, budgets, and advantages are fixed within each update.

The local branch averages over tokens within each action and then sums across actions, so $\delta_{i,t}$ is neither diluted by action length nor attenuated by clipping of length-rescaled token advantages.
At the behavior policy, the split objective has the same gradient as the additive token advantage
\begin{equation}
 A_i^{\mathrm{out}}+
 L_i\sum_{t:B_{i,t}>0}\frac{\delta_{i,t}}{B_{i,t}}m_{i,t,k}.
\label{eq:additive}
\end{equation}
Once ratio clipping activates, opposite-sign outcome and local advantages may fall into different clipping branches, so the split objective can differ from optimizing their combined advantage.

Source funding also bounds the local surrogate.
For a fixed audited trajectory whose positive events are all self-funded, let
$N_i^-=\sum_t\max\{-\delta_{i,t},0\}$.
Then
\begin{equation}
 J_i^{\mathrm{loc}}\leq
 [(1+\epsilon)\eta-(1-\epsilon)]N_i^-\leq0
\label{eq:loop}
\end{equation}
whenever $\eta\leq(1-\epsilon)/(1+\epsilon)$ and $0<\epsilon<1$.
Under these conditions, repairing self-induced errors cannot yield a net positive local surrogate.
This fixed-batch guarantee depends on correct attribution and does not by itself constrain learned behavior under shared policy parameters (\S\ref{app:escrow}).
Optimization and cost details are given in \S\ref{app:optimization}.

\begin{table*}[t]
\centering
\small
\setlength{\tabcolsep}{3pt}
\renewcommand{\arraystretch}{1.15}
\setlength{\doublerulesep}{1.5pt}
\setlength{\aboverulesep}{0pt}
\setlength{\belowrulesep}{0pt}
\begin{tabularx}{\textwidth}{c|l|*{7}{>{\centering\arraybackslash}X}}
\hline\hline
\multicolumn{2}{c|}{} & \multicolumn{3}{c}{\textbf{$\tau$-bench}} & \multicolumn{4}{c}{\textbf{$\tau^2$-Bench}} \\[1pt]
\cmidrule(lr){3-5}\cmidrule(l){6-9}
\multicolumn{2}{c|}{\textbf{Model}} & \rule{0pt}{2.6ex}\textbf{Retail} & \textbf{Airline} & \textbf{Overall} & \textbf{Retail} & \textbf{Airline} & \textbf{Telecom} & \textbf{Overall} \\
\hline
\multirow{4}{*}{4B} & Qwen3~\citep{yang2025qwen3} & 40.87 & 26.00 & 33.44 & 19.96 & \underline{21.00} & 9.43 & 15.83 \\
 & EnvScaler~\citep{song2026envscaler} & \underline{49.57} & \underline{48.00} & \underline{48.78} & \underline{50.00} & \textbf{36.00} & 13.16 & \underline{33.05} \\
 & Arctic-AWM~\citep{wang2026awm} & 30.15 & 23.00 & 26.58 & 30.26 & 19.00 & \underline{16.45} & 22.57 \\
\rowcolor{lightpurplerow}
\cellcolor{white} & \textbf{ECHO (ours)} & \textbf{52.03} & \textbf{49.50} & \textbf{50.77} & \textbf{54.09} & \textbf{36.00} & \textbf{17.33} & \textbf{35.81} \\
\hline
\multirow{7}{*}{8B} & Qwen3~\citep{yang2025qwen3} & 45.79 & 27.50 & 36.65 & 34.43 & 26.50 & 18.42 & 26.44 \\
 & EnvScaler~\citep{song2026envscaler} & 50.43 & 38.00 & 44.22 & \underline{53.51} & 36.00 & 15.79 & 35.10 \\
 & Arctic-AWM~\citep{wang2026awm} & 39.88 & 35.40 & 37.64 & 41.23 & 38.50 & \underline{23.47} & 33.45 \\
 & Klear-AgentForge~\citep{wang2025klearagentforge} & \underline{56.70} & \underline{41.50} & \underline{49.10} & 49.42 & 36.67 & 21.64 & 35.91 \\
 & MUA-RL~\citep{zhao2025muarl} & 56.50 & 29.50 & 43.00 & 49.80 & 19.00 & 21.80 & 30.20 \\
 & Simia-RL~\citep{li2025simia} & 51.82 & 36.71 & 44.27 & 52.90 & \textbf{49.00} & 18.58 & \underline{40.16} \\
\rowcolor{lightpurplerow}
\cellcolor{white} & \textbf{ECHO (ours)} & \textbf{58.17} & \textbf{43.00} & \textbf{50.59} & \textbf{54.63} & \underline{42.10} & \textbf{24.95} & \textbf{40.56} \\
\hline\hline
\end{tabularx}
\vspace{-0.3cm}
\caption{
Domain-level results on $\tau$-bench and $\tau^2$-Bench at the 4B and 8B scales.
All scores are reported as percentages.
\textbf{Bold} and \underline{underlined} values mark the best and second-best results within each model scale.
}
\vspace{-0.2cm}
\label{tab:domain-results}
\end{table*}

\begin{table*}[ht]
\centering
\small
\setlength{\tabcolsep}{3pt}
\renewcommand{\arraystretch}{1.15}
\setlength{\doublerulesep}{1.5pt}
\renewcommand\tabularxcolumn[1]{m{#1}}
\begin{tabularx}{\textwidth}{c|l|*{5}{>{\centering\arraybackslash}X}}
\hline\hline
\multicolumn{2}{c|}{\textbf{Model}} & \textbf{Base} & \textbf{Missing Function} & \textbf{Missing Parameter} & \textbf{Long Context} & \textbf{Overall} \\
\hline
\multirow{4}{*}{4B} & Qwen3~\citep{yang2025qwen3} & 32.00 & 20.00 & 24.00 & 25.50 & 25.38 \\
 & EnvScaler~\citep{song2026envscaler} & 46.50 & \underline{38.50} & 28.50 & \underline{35.50} & 37.25 \\
 & Arctic-AWM~\citep{wang2026awm} & \underline{48.55} & 36.50 & \textbf{34.22} & \textbf{36.30} & \underline{38.89} \\
\rowcolor{lightpurplerow}
\cellcolor{white} & \textbf{ECHO (ours)} & \textbf{50.50} & \textbf{46.58} & \underline{32.59} & 35.27 & \textbf{41.24} \\
\hline
\multirow{7}{*}{8B} & Qwen3~\citep{yang2025qwen3} & 32.00 & 33.50 & 22.00 & 28.00 & 28.88 \\
 & EnvScaler~\citep{song2026envscaler} & 50.50 & 41.50 & 29.50 & \underline{44.00} & 41.38 \\
 & Arctic-AWM~\citep{wang2026awm} & 53.40 & \underline{45.00} & 32.40 & \underline{44.00} & \underline{43.70} \\
 & Klear-AgentForge~\citep{wang2025klearagentforge} & 45.86 & 39.00 & 34.50 & 32.00 & 37.84 \\
 & MUA-RL~\citep{zhao2025muarl} & 21.00 & 11.50 & 15.00 & 11.00 & 14.63 \\
 & Simia-RL~\citep{li2025simia} & \underline{54.50} & 36.50 & \underline{37.54} & 38.00 & 41.64 \\
\rowcolor{lightpurplerow}
\cellcolor{white} & \textbf{ECHO (ours)} & \textbf{59.50} & \textbf{51.50} & \textbf{40.68} & \textbf{45.00} & \textbf{49.17} \\
\hline\hline
\end{tabularx}
\vspace{-0.3cm}
\caption{
BFCL-v3 Multi-Turn results across Base, Missing Function, Missing Parameter, and Long Context conditions at the 4B and 8B scales.
Overall is the unweighted mean across the four conditions.
All scores are reported as percentages.
\textbf{Bold} and \underline{underlined} values mark the best and second-best results within each model scale.
}
\vspace{-0.2cm}
\label{tab:bfcl-results}
\end{table*}

\section{Experiments}
\label{sec:experiments}

We evaluate ECHO's overall performance across six tool-use benchmarks at the 4B and 8B scales (\S\ref{sec:main-results}), illustrate responsibility-related behavioral differences between Qwen3 and ECHO through case studies (\S\ref{sec:case-study}), examine its components and reward design through ablations (\S\ref{sec:component-ablation}, \S\ref{sec:reward-ablation}), and analyze the effects of frozen-judge choice (\S\ref{sec:judge-analysis}).

\subsection{Setup}
\label{sec:experimental-setup}

\noindent
\textbf{Benchmarks and metrics.}
We evaluate ECHO at the 4B and 8B scales on BFCL-v3 Multi-Turn, BFCL-v4, $\tau$-bench, $\tau^2$-Bench, ToolSandbox, and ACEBench-Agent.
BFCL evaluates function calling, with v4 additionally covering web search and memory use \citep{patil2025bfcl,mao2025bfclv4}.
We evaluate both $\tau$-bench domains, retail and airline \citep{yao2025taubench}, and all three $\tau^2$-Bench domains, retail, airline, and telecom \citep{barres2025tau2}.
These conversational benchmarks test policy-constrained task completion, with telecom additionally requiring coordinated agent--user actions.
ToolSandbox evaluates intermediate and final state milestones \citep{lu2025toolsandbox}, while ACEBench-Agent covers multi-step and multi-turn tool use in English and Chinese \citep{chen2025acebench}.

\noindent
\textbf{Baselines.}
Qwen3-4B and Qwen3-8B serve as the base policies at 4B and 8B, respectively \citep{yang2025qwen3}.
At 4B, we compare against EnvScaler \citep{song2026envscaler} and Arctic-AWM \citep{wang2026awm}.
At 8B, we additionally compare against Klear-AgentForge \citep{wang2025klearagentforge}, MUA-RL \citep{zhao2025muarl}, and Simia-RL \citep{li2025simia}.

\noindent
\textbf{Training and configuration.}
ECHO is trained from Qwen3-4B and Qwen3-8B on the $\tau^2$-Bench training split.
Training details and hyperparameters are given in \S\ref{app:implementation}.
The frozen-judge protocol and reliability gating are described in \S\ref{app:audit} and \S\ref{app:calibration}.

\subsection{Main Results}
\label{sec:main-results}
\noindent
\textbf{Overall performance.}
ECHO achieves the best result on five of six benchmarks at both model scales.
At 4B, it outperforms the strongest same-scale baseline by 2.35 points on BFCL-v3 Multi-Turn, 0.79 on BFCL-v4, 1.99 on $\tau$-bench, 2.76 on $\tau^2$-Bench, and 8.12 on ToolSandbox.
At 8B, the corresponding gains are 5.47, 1.19, 1.49, 0.40, and 5.57 points.
ACEBench-Agent is the only benchmark on which ECHO does not lead, trailing the best baseline by 0.87 points at 4B and 0.62 at 8B.
The largest gain at both scales is on ToolSandbox, which is not used during training, suggesting that ECHO's improvements transfer beyond the training environment.

\noindent
\textbf{Domain-level results.}
At 4B, ECHO leads or ties the strongest baseline across all five domains.
At 8B, it leads $\tau$-bench Retail and Airline by 1.47 and 1.50 points, and $\tau^2$-Bench Retail and Telecom by 1.12 and 1.48 points, respectively.
The only exception is $\tau^2$-Bench Airline, where ECHO scores 42.10 versus 49.00 for Simia-RL.
ECHO nevertheless achieves the highest $\tau^2$-Bench overall score, 40.56 versus 40.16 for Simia-RL, as its 6.37-point Telecom and 1.73-point Retail advantages over Simia-RL offset the 6.90-point Airline deficit.

\noindent
\textbf{BFCL-v3 condition-level results.}
At 4B, ECHO leads on Base and Missing Function by 1.95 and 8.08 points, respectively, but trails the strongest baseline by 1.63 points on Missing Parameter and 1.03 on Long Context.
At 8B, ECHO leads all four conditions, with gains of 5.00 points on Base, 6.50 on Missing Function, 3.14 on Missing Parameter, and 1.00 on Long Context.
This advantage extends to all four conditions at 8B, whereas at 4B it is concentrated in Base and Missing Function.

\subsection{Case Studies}
\label{sec:case-study}

\begin{figure*}[t]
\centering
\includegraphics[width=\textwidth]{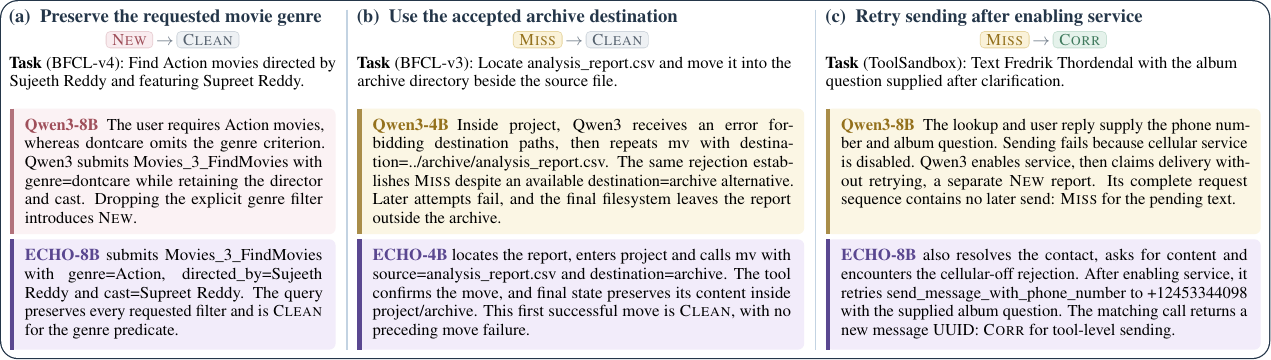}
\par\vspace{-0.2cm}
\caption{Case studies comparing agent behavior before (Qwen3) and after (ECHO) training on the same task.}
\par\vspace{-0.1cm}
\label{fig:case-study}
\end{figure*}

Figure~\ref{fig:case-study} compares Qwen3 and ECHO on matched task requirements; \textsc{Clean} indicates no responsibility event for the assessed requirement (\S\ref{app:decision-rule}).
In (a), Qwen3 replaces the requested Action genre with \texttt{dontcare}, a \textsc{New} error, while ECHO preserves all filters.
In (b), Qwen3 repeats a path already rejected by \texttt{mv}, missing an available correction, while ECHO uses the accepted directory name directly.
In (c), after a failed send, both models enable cellular service, but only ECHO resends and receives a new message identifier, whereas Qwen3 claims delivery without resending. These cases reflect ECHO's supervision: penalizing \textsc{New} and \textsc{Miss} while crediting eligible, verified corrections (\S\ref{sec:budgets}). Additional cases are provided in \S\ref{app:case-study}.

\subsection{Component Ablation}
\label{sec:component-ablation}

\begin{figure}[htbp]
  \centering
  \includegraphics[width=\columnwidth]{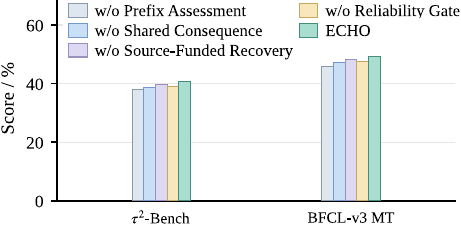}
  \vspace{-0.7cm}
  \caption{Ablation of ECHO components on $\tau^2$-Bench and BFCL-v3 Multi-Turn at the 8B scale.}
  \par\vspace{-0.4cm}
  \label{fig:component-ablation}
\end{figure}

Figure~\ref{fig:component-ablation} presents one-component-at-a-time ablations of ECHO.
w/o Prefix Assessment omits independent prefix assessment;
w/o Shared Consequence assigns full consequence mass to each event;
w/o Source-Funded Recovery removes source-funded recovery constraints; and
w/o Reliability Gate disables calibrated semantic filtering.
All four ablations reduce performance on both benchmarks, with a consistent ordering of drops.
Removing prefix assessment causes the largest degradation, suggesting a central role for decision-time responsibility, whereas removing source-funded recovery has the smallest effect.

\subsection{Reward Ablation}
\label{sec:reward-ablation}

\begin{figure}[htbp]
  \centering
  \includegraphics[width=\columnwidth]{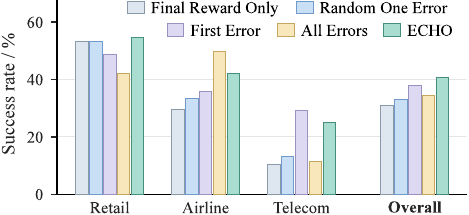}
  \vspace{-0.7cm}
  \caption{Reward ablation on $\tau^2$-Bench at the 8B scale. Each group compares five supervision configurations, and Overall denotes the mean across the three domains.}
  \label{fig:reward-tau2}
  \vspace{-0.2cm}
\end{figure}

At the 8B scale, we compare ECHO with Final Reward Only, Random One Error, First Error, and All Errors.
Final Reward Only retains terminal supervision, while the three heuristics penalize one random error, the first error, or all identified errors, respectively.
ECHO exceeds Final Reward Only by 9.51 points and First Error, the strongest heuristic, by 2.77 points overall.
Although All Errors and First Error lead Airline and Telecom, respectively, ECHO achieves the highest Retail score and the highest overall score.
These results suggest that ECHO benefits not simply from additional local supervision, but from responsibility-aware credit assignment across errors and corrections.
\S\ref{app:reward-ablation} reports the same comparison on BFCL-v3 Multi-Turn.

\subsection{Effect of the Frozen Judge}
\label{sec:judge-analysis}

\begin{figure}[htbp]
\centering
\includegraphics[width=\columnwidth]{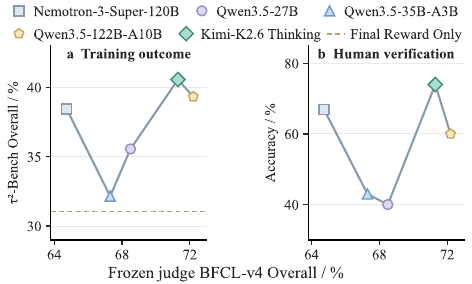}
\vspace{-0.7cm}
\caption{
Frozen-judge comparison at the 8B scale.
Left: downstream ECHO performance on $\tau^2$-Bench.
Right: responsibility identification accuracy on 100 manually verified samples before reliability gating.
}
\vspace{-0.1cm}
\label{fig:judge-analysis}
\end{figure}

Frozen-judge comparisons examine whether general tool-use capability predicts effectiveness as a source of local supervision.
Figure~\ref{fig:judge-analysis} compares five judges by BFCL-v4 score, downstream ECHO-8B performance, and responsibility identification accuracy on 100 manually verified samples.
Nemotron-3-Super-120B has the lowest BFCL-v4 score, yet yields stronger downstream ECHO performance than both smaller Qwen judges.
Kimi-K2.6 Thinking produces the strongest ECHO policy, exceeding Qwen3.5-122B-A10B by 1.24 percentage points despite the latter's higher BFCL-v4 score.
These reversals suggest that general tool-use capability is an incomplete proxy for supervision effectiveness.

\section{Conclusion}
We presented ECHO, a responsibility-aware credit assignment framework for tool-use agents that distinguishes newly introduced errors, inherited consequences, missed corrections, and successful corrections, and converts them into bounded action-level supervision. This design addresses three limitations of conventional terminal or step-level feedback: (i) conflating error manifestations with responsible decisions, (ii) double-counting shared downstream consequences, and (iii) over-crediting recovery from self-induced errors. Across six tool-use benchmarks and two model scales, ECHO achieves strong and broadly consistent improvements over the compared baselines. Together, these results support decision responsibility as a useful perspective for fine-grained credit assignment across complex tool-use trajectories.

\section*{Limitations}

This work focuses specifically on responsibility-aware credit assignment for tool-use agents. (i) ECHO is designed for training-time credit assignment rather than inference-time intervention. Techniques such as reflection, replanning, and online verification are complementary to the proposed framework and are not considered here. (ii) Our analysis centers on responsibility attribution within agent--tool interaction trajectories. Broader system-level considerations, including tool construction, multi-agent coordination, and interface design, fall outside the scope of the proposed framework.

\section*{Ethical Considerations}
All experiments in this study were conducted using publicly available datasets and models licensed for research use. To the best of our knowledge, we do not identify any method-specific ethical concerns beyond those generally associated with tool-use agents and automated model evaluation.

\bibliography{custom}

\appendix
\startcontents[appendices]
\section*{Contents of Appendix}
\printcontents[appendices]{}{1}{\normalsize}

\section{Audit Protocol and Evidence Handling}
\label{app:audit}

\subsection{Candidate Selection and Independent Requests}

The full-trajectory proposer returns at most sixteen event hypotheses and explicitly flags any it omits.
Each hypothesis identifies its action, policy span, versioned predicate, deviation origin, evidence pointers, and proposed responsibility type.
Impact records also specify propagation links, severity, first-restoration evidence, and authenticated origins.
Validated fields determine training weights through the scoring rules in \S\ref{sec:accounting}--\ref{sec:budgets}.

The prefix-review budget covers $K=8$ distinct actions.
Up to six proposed actions are selected in chronological order, and two non-proposed sentinel actions are selected by a fixed-seed hash ordering.
The same ordering fills unused proposal slots until the budget or the available actions are exhausted.
Each prefix request returns at most four predicate-level records, and omitted records abstain.

Each prefix request contains the recorded decision-time context $h_{i,t}$, the current action, and the available tool contracts.
Retrospective labels, selected predicates, rationales, severity, trajectory suffixes, and hidden state are excluded from these inputs.
Batched requests keep isolated attention contexts, and both judge roles use the same frozen checkpoint with fixed prompts and decoding.
Prefix assessment returns supported defects, legitimate repair intentions, defect-free assessments, or abstention.
A successful correction is established only when an accepted repair intention is paired with execution confirmation.

Defects found at sentinel actions contribute only through the direct-defect term of Eq.~\eqref{eq:loss}, provided their prefix evidence passes validation.
Unsupported impact and propagation terms receive zero weight, and sentinel repairs receive positive supervision only with authenticated restoration evidence.

\subsection{Responsibility Boundaries}

\paragraph{Inherited inputs and current obligations.}
The classification of a reused product identifier depends on the discrepancy evidence visible when it is used.
A visible mismatch and an available resolving lookup can establish a missed correction when the agent continues committing to the wrong product.
An apparently valid opaque identifier can carry an inherited consequence when its defect becomes observable later.
Ambiguous cases receive local abstention, preserving the distinction between available evidence and retrospectively established error.

\paragraph{Feasible responses.}
The feasible alternative response (witness) of a missed correction specifies a concrete tool or response, known arguments, availability under the visible contract, and the discrepancy addressed.
Clarification qualifies when the user can disambiguate exposed alternatives; stopping qualifies when permitted and sufficient to avoid the identified impermissible operation.
A legitimate repair attempt retains its prefix-grounded assessment even if an unpredictable outage later prevents restoration.

\paragraph{Mixed actions and verified restoration.}
An action that restores one field while violating another produces distinct positive and negative events.
Transaction identifiers or field-level provenance link verified restoration to the responsible action.
Ambiguous restoration attribution triggers abstention from positive supervision.
A verified correction remains valid despite a separate terminal failure, with auxiliary eligibility determined by its origin and available source funding.

\paragraph{Visibility and external origins.}
Decision-time visibility follows delivery timestamps and actual context inclusion, placing synchronous tool output after its generating action.
Hidden snapshots support retrospective impairment assessment, while correction opportunities require discrepancy evidence visible in $h_{i,t}$.
Initial-state records authenticate pre-existing defects, and environment records authenticate subsequent external incidents.
Omitted sources and unsupported policy claims leave origins unknown.
A new policy error that reopens a repaired predicate creates a separate self-induced origin with its own source penalty.

\section{Calibration of Selective Supervision}
\label{app:calibration}

Prompt development, gate calibration, hyperparameter selection, and final evaluation use disjoint task partitions.
A fixed selection rule samples one eligible event per independent task instance within each predeclared evidence stratum.
Events and rollouts sharing a task form one sampling cluster.
Two annotators assess prefix-grounded responsibility, followed by a separate retrospective review of impact, provenance, and restoration.
Adjudication is blinded to training coefficients, and correctness requires validity of every training-relevant semantic field.

For stratum $b$, let $k_b$ of $n_b$ independent calibration units contain an incorrect assertion.
With $B_{\mathrm{cal}}$ predeclared strata and simultaneous failure level $\delta_{\mathrm{cal}}$, define
\begin{equation}
 U_b=\operatorname{Beta}^{-1}
 \left[1-\frac{\delta_{\mathrm{cal}}}{B_{\mathrm{cal}}};
 k_b+1,n_b-k_b\right]
\label{eq:calibration-upper}
\end{equation}
for $0\leq k_b<n_b$, and set $U_b=1$ if $k_b=n_b$ or $n_b=0$.
The inverse beta distribution yields a one-sided exact binomial upper confidence limit.
Combining it with the indicator $V_e$ of the structural checks in \S\ref{sec:assessment} gives
\begin{equation}
 g_e=V_e\mathbf1\{n_{b(e)}\geq n_{\min}\}
 \mathbf1\{U_{b(e)}\leq\epsilon_J\}.
\label{eq:gate}
\end{equation}
Independent sampling from the same stratum distributions gives simultaneous upper coverage of at least $1-\delta_{\mathrm{cal}}$ through a union bound.
This guarantee concerns stratum-level semantic-error rates under the calibration distribution.
Individual-event validity, impact magnitude, and reliability under policy shift remain separate assessment targets.
Post hoc stratum selection uses an independent selection split to preserve the calibration protocol.

The calibration design crosses three supervised responsibility types with explicit or semantic evidence to form six strata.
Thresholds are $\delta_{\mathrm{cal}}=0.05$, $\epsilon_J=0.10$, and $n_{\min}=50$; insufficiently sampled or unreliable strata abstain.
Precision and coverage jointly characterize selective supervision, while monitoring newly annotated training traces supports drift detection and recalibration on fresh development data.

\section{Derivations and Guarantee Boundaries}
\label{app:proofs}

\subsection{Shared-Consequence and Budget Bounds}

Equation~\eqref{eq:ownership} gives $\sum_ea_{ec}=1-\zeta_c\leq1$ for nonempty $P_c$, and zero otherwise.
Let $v_e^{\mathrm{shared}}$ denote the shared-consequence component of a negative event's weight.
Since $g_e\leq1$ and $p_e\leq1$,
\begin{align}
 \sum_ev_e^{\mathrm{shared}}
 &=(1-\kappa)\sum_c M_c
   \sum_e a_{ec}g_e[1+\alpha p_e]\nonumber\\
 &\leq(1-\kappa)(1+\alpha)\sum_cM_c.
\label{eq:shared-bound}
\end{align}
The shared-consequence bound is independent of the number of implicated events; direct-defect terms separately supervise certified decisions.
Duplicating observations preserves the accounting when validated predicates, event keys, provenance links, and first-restoration sets are fixed.
The invariance therefore concerns validated records, with proposer sensitivity to redundant text requiring empirical assessment.

Equation~\eqref{eq:budget} caps each directional total at $\lambda_s$.
Source funding in Eq.~\eqref{eq:escrow} only reduces or preserves positive budgets, giving
\begin{equation}
 \sum_t|\delta_{i,t}|
 \leq\sum_e b_e^-+\sum_r\widetilde b_r^+
 \leq\lambda_-+\lambda_+.
\end{equation}
The resulting bound limits total auxiliary budget, while additional responsibility events can reduce individual normalized shares.

\subsection{First-Order Equivalence}

At the behavior policy, likelihood ratios equal one and lie inside the clipping interval.
Writing the token score gradient as $s_{i,k}=\nabla_\theta\log\pi_\theta[y_{i,k}\mid h_{i,k}]$, where $h_{i,k}$ is the token-level context, the uncapped additive-token surrogate gives
\begin{align}
 \nabla J_i^{\mathrm{add}}
 &=\frac{A_i^{\mathrm{out}}}{L_i}\sum_kq_{i,k}s_{i,k}
 \nonumber\\
 &\quad+\sum_{t:B_{i,t}>0}\frac{\delta_{i,t}}{B_{i,t}}
    \sum_km_{i,t,k}s_{i,k}\nonumber\\
 &=\nabla[J_i^{\mathrm{out}}+J_i^{\mathrm{loc}}].
\end{align}
Objective values also agree whenever all relevant ratios lie strictly inside the clipping interval.
Outside it, opposite-sign advantages can activate different clipping branches, yielding distinct objectives with an identical KL term.

If each token score gradient has norm at most $G_0$ at the behavior policy, then
\begin{equation}
 \|\nabla J_i^{\mathrm{loc}}\|
 \leq G_0\sum_t|\delta_{i,t}|
 \leq G_0[\lambda_-+\lambda_+].
\end{equation}
This gradient bound holds under the score-gradient assumption at the sampling policy.
Extending it across tokenizations or away from that policy requires additional assumptions.

\subsection{Source Funding and the Clipped Local Surrogate}
\label{app:escrow}

For a fixed audited trajectory with exclusively self-funded positive events, define gross budgets $P_0=\sum_r\widetilde b_r^+$ and $N_0=\sum_e b_e^-$.
Each source penalty $D_d$ in Eq.~\eqref{eq:principal} sums a disjoint subset of negative charges, and paid repairs of origin $d$ total at most $\eta D_d$, yielding
\begin{equation}
 P_0\leq\eta\sum_dD_d\leq\eta N_0.
\end{equation}
Let $C$ be the amount canceled when positive and negative events share actions.
Net masses $P=P_0-C$ and $N=N_0-C$, where $N=N_i^-$ as defined in Eq.~\eqref{eq:loop}, satisfy
\begin{equation}
 P-\eta N=P_0-\eta N_0-(1-\eta)C\leq0
\end{equation}
for $0\leq\eta\leq1$.
For any $\rho>0$ and $a,b\geq0$, the clipped surrogate obeys
\begin{align}
 \phi_\epsilon(\rho,a)&\leq(1+\epsilon)a,\\
 \phi_\epsilon(\rho,-b)&\leq-(1-\epsilon)b.
\end{align}
Averaging over each action's tokens preserves these inequalities, yielding
\begin{align}
 J_i^{\mathrm{loc}}
 &\leq(1+\epsilon)P-(1-\epsilon)N\nonumber\\
 &\leq[(1+\epsilon)\eta-(1-\epsilon)]N.
\end{align}
This establishes Eq.~\eqref{eq:loop} under its stated condition.
Within trajectories containing external bonuses, the bound applies to self-induced components with action-disjoint support or independently evaluated local surrogates.
These conditions preserve component-level accounting under nonlinear clipping.

Shared policy parameters can increase error creation despite the scalar source-funding bound.
Consider an erroneous action with probability $p[\theta]=\sigma[\theta]$ and a subsequent repair with probability $q[\theta]=\sigma[H\theta]$.
Assign coefficients $-D$ to the source and $\eta D$ to its verified repair.
At $\theta=0$, the expected local score-gradient update is
\begin{equation}
 g_{\mathrm{loc}}[0]=-\frac D4+\frac{\eta D H}{8}.
\end{equation}
For $H>2/\eta$, the update increases error probability while satisfying the source-funding constraint.
The counterexample separates a bounded local surrogate from behavioral guarantees and motivates empirical assessment of error introduction, correction, and repeated repair cycles.

\subsection{Observational Attribution Limits}

\label{app:attribution-limits}

A failed trajectory can support causal models with different intervention effects.
For binary errors $E_1,E_2$, both $F=E_1\lor E_2$ and $F=E_1\land E_2$ explain $E_1=E_2=F=1$.
Correcting $E_1$ preserves failure under the first model and resolves it under the second.
The trace therefore leaves intervention-based responsibility underdetermined.
ECHO uses the ownership convention in Eq.~\eqref{eq:ownership} to allocate observed constraint impairment.

\section{Implementation Details}
\label{app:implementation}

\subsection{Training Data and Implementation}

ECHO is initialized from Qwen3-4B and Qwen3-8B and trained through reinforcement learning.
We use 92 tasks from the $\tau^2$-Bench training splits, comprising 66 retail and 26 airline tasks.
A further 8 retail and 4 airline tasks form the 12-task development split.
Both scales use 75 updates, 12 tasks per batch, eight rollouts per task, and a learning rate of $5\times10^{-6}$.

\subsection{ECHO Configuration}

Kimi-K2.6 Thinking serves as the fixed judge for retrospective proposals and independent prefix assessments.
Each trajectory receives one proposal request and at most eight prefix assessments, with six proposal slots and two sentinel slots.
We use $\kappa=0.25$, $\alpha=0.5$, $\lambda_-=\lambda_+=0.5$, $\eta=0.5$, and clipping half-width $\epsilon=0.2$.
Outcome advantages use population-standard-deviation normalization with $\varepsilon=10^{-6}$.
\S\ref{app:audit} and \S\ref{app:calibration} specify evidence validation and reliability gating.

\section{Optimization Details and Computational Cost}
\label{app:optimization}

\subsection{Update Sequence}

Algorithm~\ref{alg:echo} summarizes one update.
Each update audits successful and failed trajectories and retains every group reward for outcome normalization.
Negative budgets determine source penalties before eligible corrections are normalized and capped by source funding.
Aggregated action coefficients remain fixed during optimization of Eq.~\eqref{eq:objective}; the judge operates during training.
The coefficients in \S\ref{app:implementation} satisfy the source-funding condition $\eta\leq[1-\epsilon]/[1+\epsilon]$.

Stored behavior probabilities correspond to the rollout distribution, including its temperature and token support.
The sampling convention uses temperature one with full token support, and distributed accumulation averages over contributing trajectories across workers.

\subsection{Audit Cost}

Auditing a trajectory uses one proposal request and at most $K$ independent prefix requests.
For input lengths $n$ and output lengths $m$, the token cost is
\begin{align}
 C_i^{\mathrm{judge}}
 &=n_i^{\mathrm{full}}+m_i^{\mathrm{full}}
 \nonumber\\
 &\quad+\sum_{t\in S_i}
 [n_{i,t}^{\mathrm{pre}}+m_{i,t}^{\mathrm{pre}}]
 +C_i^{\mathrm{retry}}.
\end{align}
Requirement-inventory compilation is amortized per task and accounted for separately.
Uncached prefix input scales up to $O(Kn_i)$; exact-prefix caching can reduce repeated computation, while batching preserves request count.
Inference cost also depends on context length and attention implementation.

\subsection{Realized Local Coefficients}

Away from clipping boundaries, an active local token contributes $\delta_{i,t}\rho_{i,k}/B_{i,t}$ to its score-gradient coefficient.
Define
\begin{align}
 \chi_{i,t,k}
 &=\mathbf1\{\delta_{i,t}>0,\rho_{i,k}\leq1+\epsilon\}
 \nonumber\\
 &\quad+\mathbf1\{\delta_{i,t}<0,\rho_{i,k}\geq1-\epsilon\},\\
 \delta_{i,t}^{\mathrm{eff}}
 &=\frac{\delta_{i,t}}{B_{i,t}}
 \sum_km_{i,t,k}\rho_{i,k}\chi_{i,t,k}.
\end{align}
Active negative-advantage ratios can exceed the upper clipping threshold, so realized coefficient magnitudes can exceed their on-policy masses.
Realized coefficients and sign-specific clipping fractions characterize how supervision changes through aggregation and clipping.
Gradient norms and local-to-outcome alignment further distinguish update magnitude from directional agreement.

\section{Reward Ablation on Multi-Turn Function Calling}
\label{app:reward-ablation}

\begin{figure}[htbp]
  \centering
  \includegraphics[width=\columnwidth]{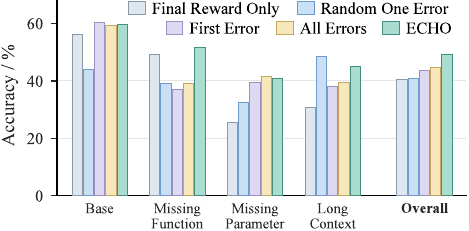}
  \caption{Reward ablation on BFCL-v3 Multi-Turn at 8B. Overall is the mean across Base, Missing Function, Missing Parameter, and Long Context.}
  \vspace{-0.2cm}
  \label{fig:reward-bfclv3}
\end{figure}

Figure~\ref{fig:reward-bfclv3} repeats the reward ablation of \S\ref{sec:reward-ablation} on BFCL-v3 Multi-Turn, which evaluates tool use under different function, parameter, and context constraints.
ECHO exceeds Final Reward Only by 8.76 percentage points and All Errors, the strongest heuristic, by 4.42 points overall.
Its advantage over All Errors is concentrated in Missing Function, where all three error-selection heuristics fall below terminal supervision.
ECHO instead improves on this subset while retaining gains over Final Reward Only in Missing Parameter and Long Context.

Subset leaders vary: First Error leads Base, All Errors leads Missing Parameter, and Random One Error leads Long Context.
Random One Error's Long Context lead, however, comes with a substantial Base deficit, illustrating the uneven effects of unstructured selection.
The strongest heuristic also changes from First Error on $\tau^2$-Bench to All Errors on BFCL-v3 Multi-Turn.
Although heuristics lead on individual subsets, ECHO achieves the highest overall score on both benchmarks.

\section{Responsibility Event Taxonomy and Benchmark Coverage}
\label{app:taxonomy}

\begin{table*}[t]
\centering
\footnotesize
\setlength{\tabcolsep}{4pt}
\renewcommand{\arraystretch}{1.08}
\begin{tabularx}{\textwidth}{@{}>{\raggedright\arraybackslash}p{0.20\textwidth}>{\raggedright\arraybackslash}p{0.29\textwidth}>{\raggedright\arraybackslash}X@{}}
\toprule
\textbf{Semantic family} & \textbf{Observed diagnostics} & \textbf{Evidence required for responsibility assignment} \\
\midrule
Format, schema, and types & Malformed calls; parameter or type mismatches. & Visible schema constraints, actionable rejections, and execution of a corrected call. \\
Tool selection and call count & Incorrect function, call count, or relevance decision. & Available tools, selection obligations, and provenance of an inherited choice. \\
Arguments, entities, and values & Incorrect argument, entity, value, or parallel-call match. & Visible value constraints, the source of reused data, and verified value correction. \\
State and prerequisites & State mismatch; unmet database, environment, or milestone checks. & Applicable preconditions, the responsible state transition, and verified restoration. \\
Execution and omitted work & Missing or mismatched execution; process-score deficits. & Required operations and current progress, with a feasible continuation or correction. \\
Retrieval, memory, and answers & Expected answer absent after search or memory use. & Retrieved evidence, stored-state provenance, and correction of the affected fact. \\
Communication obligations & Unmet communication checks. & Visible disclosure or clarification obligations and an available response. \\
Forbidden outcomes and guardrails & Forbidden-state or minefield matches. & Applicable public constraints and the action that introduced or propagated the violation. \\
Progress, stopping, and limits & Empty turns, step limits, truncation, or forced termination. & The recorded stopping decision, remaining obligations, and feasible progress. \\
Runtime and external boundary & Explicit runtime or resource errors. & Incident origin and the agent's response, with execution evidence of recovery. \\
Aggregate outcome only & Reported aggregate similarity deficits without a localized check. & Task requirement predicates (\S\ref{sec:accounting}), assessed independently of the score. \\
\bottomrule
\end{tabularx}
\caption{
Semantic families of checker diagnostics and the evidence each requires for responsibility assignment.
Diagnostics (23 literal \texttt{error\_type} codes and 19 derived signals) are drawn from 14,014 scored Qwen3 base-model trajectories (7,007 each at 4B and 8B) across the six benchmarks.
Families can overlap within a trajectory.
A further 2,319 unlocalized or unmapped occurrences fall outside these families (\S\ref{app:diagnostic-codebook}).
}
\label{tab:semantic-mapping}
\end{table*}

\begin{table*}[tp]
\centering
\footnotesize
\setlength{\tabcolsep}{3pt}
\renewcommand{\arraystretch}{1.1}
\begin{tabularx}{\textwidth}{@{}>{\raggedright\arraybackslash}p{0.52\textwidth}>{\raggedright\arraybackslash}X r@{}}
\toprule
\textbf{Exact diagnostic string} & \textbf{Semantic family} & \textbf{Occurrences} \\
\midrule
\texttt{\detokenize{agentic:answer_not_found}} & Retrieval, memory, and answers & 752 \\
\texttt{\detokenize{ast_decoder:decoder_failed}} & Format, schema, and types & 3 \\
\texttt{\detokenize{class attributes wrong}} & State and prerequisites & 53 \\
\texttt{\detokenize{context_limit_exceeded}} & Progress, stopping, and limits; Runtime and external boundary & 4 \\
\texttt{\detokenize{irrelevance_error:decoder_success}} & Tool selection and call count & 540 \\
\texttt{\detokenize{multiple_function_checker:wrong_count}} & Tool selection and call count & 47 \\
\texttt{\detokenize{parallel_function_checker_no_order:cannot_find_match}} & Tool selection and call count; Arguments, entities, and values & 77 \\
\texttt{\detokenize{parallel_function_checker_no_order:wrong_count}} & Tool selection and call count & 21 \\
\texttt{\detokenize{relevance_error:decoder_failed}} & Format, schema, and types & 3 \\
\texttt{\detokenize{simple_function_checker:missing_optional}} & Format, schema, and types & 28 \\
\texttt{\detokenize{simple_function_checker:missing_required}} & Format, schema, and types & 3 \\
\texttt{\detokenize{simple_function_checker:unexpected_param}} & Format, schema, and types & 7 \\
\texttt{\detokenize{simple_function_checker:wrong_count}} & Tool selection and call count & 4 \\
\texttt{\detokenize{simple_function_checker:wrong_func_name}} & Tool selection and call count & 40 \\
\texttt{\detokenize{type_error:java}} & Format, schema, and types & 8 \\
\texttt{\detokenize{type_error:js}} & Format, schema, and types & 4 \\
\texttt{\detokenize{type_error:nested}} & Format, schema, and types & 11 \\
\texttt{\detokenize{type_error:simple}} & Format, schema, and types & 39 \\
\texttt{\detokenize{value_error:dict_key}} & Arguments, entities, and values & 9 \\
\texttt{\detokenize{value_error:dict_value}} & Arguments, entities, and values & 2 \\
\texttt{\detokenize{value_error:list/tuple}} & Arguments, entities, and values & 15 \\
\texttt{\detokenize{value_error:others}} & Arguments, entities, and values & 114 \\
\texttt{\detokenize{value_error:string}} & Arguments, entities, and values & 309 \\
\bottomrule
\end{tabularx}
\caption{Literal \texttt{error\_type} codes recorded by benchmark checkers, with their semantic families and occurrence counts. A code mapped to two families lists both, separated by a semicolon.}
\label{tab:literal-codes}
\end{table*}

\begin{table*}[tp]
\centering
\footnotesize
\setlength{\tabcolsep}{3pt}
\renewcommand{\arraystretch}{1.1}
\begin{tabularx}{\textwidth}{@{}>{\raggedright\arraybackslash}p{0.52\textwidth}>{\raggedright\arraybackslash}X r@{}}
\toprule
\textbf{Exact diagnostic string} & \textbf{Semantic family} & \textbf{Occurrences} \\
\midrule
\texttt{\detokenize{acebench:process_score<1}} & Execution and omitted work & 53 \\
\texttt{\detokenize{audit:score<1_without_listed_diagnostic}} & Unmapped or unlocalized & 2,278 \\
\texttt{\detokenize{record:explicit_exception_or_error}} & Runtime and external boundary & 51 \\
\texttt{\detokenize{tau2:action_checks:action_match=false}} & Execution and omitted work & 897 \\
\texttt{\detokenize{tau2:communicate_checks:met=false}} & Communication obligations & 38 \\
\texttt{\detokenize{tau2:db_match=false}} & State and prerequisites & 365 \\
\texttt{\detokenize{tau2:env_assertions:met=false}} & State and prerequisites & 259 \\
\texttt{\detokenize{tau2:reward_breakdown:ACTION<1}} & Execution and omitted work & 15 \\
\texttt{\detokenize{tau2:reward_breakdown:COMMUNICATE<1}} & Communication obligations & 36 \\
\texttt{\detokenize{tau2:reward_breakdown:DB<1}} & State and prerequisites & 178 \\
\texttt{\detokenize{tau2:reward_breakdown:ENV_ASSERTION<1}} & State and prerequisites & 136 \\
\texttt{\detokenize{tau2:termination_reason:invalid_model_generation}} & Unmapped or unlocalized & 41 \\
\texttt{\detokenize{tau2:termination_reason:max_agent_steps_30}} & Progress, stopping, and limits & 25 \\
\texttt{\detokenize{toolsandbox:milestone_mapping:similarity<1}} & State and prerequisites & 1,502 \\
\texttt{\detokenize{toolsandbox:milestone_similarity<1}} & State and prerequisites & 1,074 \\
\texttt{\detokenize{toolsandbox:minefield_mapping:similarity>0}} & Forbidden outcomes and guardrails & 35 \\
\texttt{\detokenize{toolsandbox:minefield_similarity>0}} & Forbidden outcomes and guardrails & 35 \\
\texttt{\detokenize{toolsandbox:similarity<1}} & Aggregate outcome only & 1,109 \\
\texttt{\detokenize{trajectory:truncated=true}} & Progress, stopping, and limits & 13 \\
\bottomrule
\end{tabularx}
\caption{Derived diagnostic signals computed from benchmark records, with their semantic families and occurrence counts. Signals marked as unmapped or unlocalized fall outside the eleven families of Table~\ref{tab:semantic-mapping}.}
\label{tab:derived-signals}
\end{table*}

This section elaborates on the responsibility-event taxonomy introduced in \S\ref{sec:events}--\ref{sec:assessment}.
\S\ref{app:decision-rule} specifies how to assign at most one type to each candidate event, \S\ref{app:content-vs-responsibility} distinguishes responsibility from error content, and \S\ref{app:semantic-mapping} maps diagnostics from the six evaluation benchmarks to the evidence required for responsibility assignment.

\subsection{Event-Level Decision Rule}
\label{app:decision-rule}

Consider a candidate event $e=\langle t,f,d,z\rangle$ (Eq.~\eqref{eq:event}) in trajectory $i$.
Four evidence indicators determine $z$.
$\mathsf{Def}_e$ and $\mathsf{Opp}_e$ are judged from $(h_{i,t},a_{i,t})$ alone, whereas $\mathsf{Rep}_e$ and $\mathsf{Prov}_e$ additionally require validated full-trajectory evidence of restoration and propagation, respectively.
\begin{itemize}[leftmargin=*,itemsep=1pt,topsep=2pt]
  \item $\mathsf{Def}_e$: action $a_{i,t}$ commits a distinct defect on $f$ that is evident from a fact or constraint visible in $h_{i,t}$, and that does not repeat or continue a deviation on $f$ existing before $a_{i,t}$.
  \item $\mathsf{Rep}_e$: $a_{i,t}$ is a legitimate response to a visible deviation on $f$, and execution evidence confirms restoration of $f$ or of a required precondition.
  \item $\mathsf{Opp}_e$: a deviation on $f$ already exists, the discrepancy is visible in $h_{i,t}$, and a feasible permitted response (\S\ref{app:audit}) is not taken.
  \item $\mathsf{Prov}_e$: the consequence on $f$ is linked to an earlier deviation through a validated propagation link, and that deviation is not yet visible in $h_{i,t}$.
\end{itemize}
The type is then assigned by precedence:
{\small
\begin{equation}
\setlength{\arraycolsep}{3pt}
z=\begin{cases}
\textsc{New}, & \mathsf{Def}_e,\\
\textsc{Corr}, & \neg\mathsf{Def}_e\land\mathsf{Rep}_e,\\
\textsc{Miss}, & \neg\mathsf{Def}_e\land\neg\mathsf{Rep}_e\land\mathsf{Opp}_e,\\
\textsc{Inh}, & \neg\mathsf{Def}_e\land\neg\mathsf{Rep}_e\land\neg\mathsf{Opp}_e\land\mathsf{Prov}_e.
\end{cases}
\label{eq:decision-rule}
\end{equation}\par}

Because each case negates the indicators above it, the four types are mutually exclusive.
Precedence resolves cases in which several indicators are supported, whereas insufficient or conflicting evidence precludes a reliable assignment and leads to abstention (\S\ref{sec:assessment}).
A discrepancy counts as visible only if evidence of it is present in $h_{i,t}$ (\S\ref{app:audit}); missing information that the policy had no reason to expect does not count.
Abstention also applies when sufficient evidence matches no type: with neither $\mathsf{Def}_e$ nor $\mathsf{Rep}_e$, a visible discrepancy that admits no feasible response satisfies neither $\mathsf{Opp}_e$ nor $\mathsf{Prov}_e$, yet cannot be \textsc{Clean} because it is visible.
Responsibility is assessed separately for each predicate, so a single action can yield a \textsc{Corr} event by repairing one predicate and a \textsc{New} event by violating another; Eq.~\eqref{eq:delta} aggregates their coefficients.

\paragraph{Clean pairs and abstention.}
Two outcomes yield no responsibility type.
An action--predicate pair is \textsc{Clean} when the evidence is sufficient, none of the four indicators holds, and no discrepancy on $f$ is visible; the case studies in \S\ref{sec:case-study} and \S\ref{app:case-study} use this label.
Following selective classification \citep{elyaniv2010selective}, a candidate event abstains when the evidence is insufficient or conflicting, or when a visible discrepancy fits none of the four types; \S\ref{app:calibration} calibrates which evidence strata are admitted.
Both outcomes retain only terminal supervision but record different findings: \textsc{Clean} is a supported judgment that the decision raises no responsibility event for $f$, whereas abstention indicates that the type is either unresolved or inapplicable.

\paragraph{Relation to training signals.}
Under \S\ref{sec:budgets}, admitted \textsc{New} and \textsc{Miss} events receive negative budgets.
Admitted, execution-verified \textsc{Corr} events receive provisional positive budgets, capped by source funding for self-induced origins, exempt from that cap for authenticated external origins, and withheld for unknown origins.
\textsc{Inh} events receive no auxiliary budget and thus contribute no local correction.
Post-training differences on \textsc{Inh} cases can therefore reflect terminal supervision or indirect effects of local supervision applied to other events, such as upstream \textsc{New} errors; the comparisons in \S\ref{app:paired-cases} should be interpreted accordingly.

\subsection{Error Content versus Responsibility}
\label{app:content-vs-responsibility}

An incorrect argument, a policy violation, and a corrupted memory describe the \emph{content} of a failure, whereas responsibility describes how the current decision relates to that failure.
An unsupported identifier, for example, is \textsc{New} where it is introduced and \textsc{Inh} where later actions reuse it before the error becomes visible.
Once the error is visible, ignoring it despite a feasible response is \textsc{Miss}, whereas an execution-confirmed repair is \textsc{Corr}.
Because types depend on this relation, not on content, the same failure can receive different types across decisions, and the types apply across tools and domains without enumerating surface errors.

This distinction complements existing failure analyses.
MAST organizes semantic failure modes of multi-agent systems \citep{cemri2025mast}, and Who\&When identifies the responsible agent and decisive step in a failed interaction for post-hoc diagnosis \citep{zhang2025whowhen}; ECHO instead records each decision's role in the error-and-recovery chain and uses it for training-time credit.
ProcessBench targets the earliest erroneous step because later steps can be locally valid yet build on an incorrect premise \citep{zheng2025processbench}; ECHO likewise separates an error's origin (\textsc{New}) from its downstream consequences (\textsc{Inh}).
Reflexion and CRITIC use verbal or tool-grounded feedback to revise behavior at inference time \citep{shinn2023reflexion,gou2024critic}; ECHO instead evaluates the agent's responses to visible discrepancies for training-time credit assignment, distinguishing missed corrections (\textsc{Miss}) from execution-confirmed repairs (\textsc{Corr}).
As in step-level process supervision \citep{lightman2024verify}, each decision is assessed individually, but ECHO judges local responsibility only from the policy-visible prefix and reserves full-trajectory review for proposing candidate events and verifying propagation and restoration.

\subsection{Mapping Benchmark Diagnostics to Responsibility Evidence}
\label{app:semantic-mapping}

The six benchmarks cover complementary parts of the error chain.
BFCL's tool contracts, multi-turn tasks, and memory settings expose argument defects, unresolved prerequisites, and inherited information loss \citep{patil2025bfcl,mao2025bfclv4}.
ToolSandbox adds state dependencies and intermediate milestones \citep{lu2025toolsandbox}, $\tau$-bench domain policies and database-state evaluation \citep{yao2025taubench}, $\tau^2$-Bench user-controlled state changes \citep{barres2025tau2}, and ACEBench-Agent multi-step service interactions \citep{chen2025acebench}.

Table~\ref{tab:semantic-mapping} groups the mapped checker diagnostics recorded on Qwen3 base-model trajectories into eleven semantic families and lists the evidence required for responsibility assignment within each family.
Checker diagnostics identify failed checks but do not establish responsibility, which additionally requires the decision-time, provenance, feasibility, and restoration evidence listed in the third column.
When only an aggregate score is available, responsibility is instead assessed against task requirement predicates extracted from public instructions and tool contracts independently of outcomes (\S\ref{sec:accounting}).
The mapping serves only this analysis and is not provided to the training-time judge.

\section{Diagnostic Codebook}
\label{app:diagnostic-codebook}

Tables~\ref{tab:literal-codes} and~\ref{tab:derived-signals} list all 23 literal checker codes and 19 derived signals observed in the 14,014 scored Qwen3 base-model trajectories summarized in Table~\ref{tab:semantic-mapping}.
Literal codes are \texttt{error\_type} strings emitted by benchmark checkers, whereas derived signals are computed from benchmark records such as reward breakdowns, termination reasons, and milestone similarities.
Because not every benchmark emits literal codes, Table~\ref{tab:derived-signals} supplements them with benchmark-specific derived signals and generic \texttt{audit}, \texttt{record}, and \texttt{trajectory} signals.
Counts refer to diagnostic occurrences and are not additive: a trajectory can carry several diagnostics, and some derived signals are computed from one record, such as the database match and its reward component.

Family names follow Table~\ref{tab:semantic-mapping}, except for the unmapped or unlocalized label, which covers score deficits without a listed diagnostic and unmapped terminations.
Benchmark-reported aggregate similarity deficits form the ``Aggregate outcome only'' family.
Semantic families describe only diagnostic content, so assigning a responsibility type also requires the evidence in \S\ref{app:decision-rule}.

\renewcommand{\dbltopfraction}{0.95}
\renewcommand{\dblfloatpagefraction}{0.5}
\makeatletter
\setlength{\@dblfptop}{0pt}
\setlength{\@dblfpsep}{14pt plus 2pt}
\setlength{\@dblfpbot}{0pt plus 1fil}
\makeatother
\section{Additional Case Studies}
\label{app:case-study}

This section presents trajectory-level examples of responsibility events and behavioral differences before and after ECHO training, from evaluation runs of Qwen3 and ECHO at 4B and 8B on the benchmarks in \S\ref{sec:experimental-setup}.
Labels are assigned by applying the decision rule in \S\ref{app:decision-rule} to the displayed evidence, with information available at decision time kept separate from evidence observed later.
Single-turn BFCL items record a submitted call without execution, so their defects are established from the visible constraint and the submitted call.
In all other panels, execution claims rely on the displayed tool results or state changes.
Because BFCL-v3 and BFCL-v4 share multi-turn task categories, each panel identifies the benchmark run that produced its trajectory.
The panels are qualitative examples and do not estimate pattern frequencies.

\subsection{Responsibility Events in Base-Model Trajectories}
\label{app:base-cases}

\begin{figure*}[p]
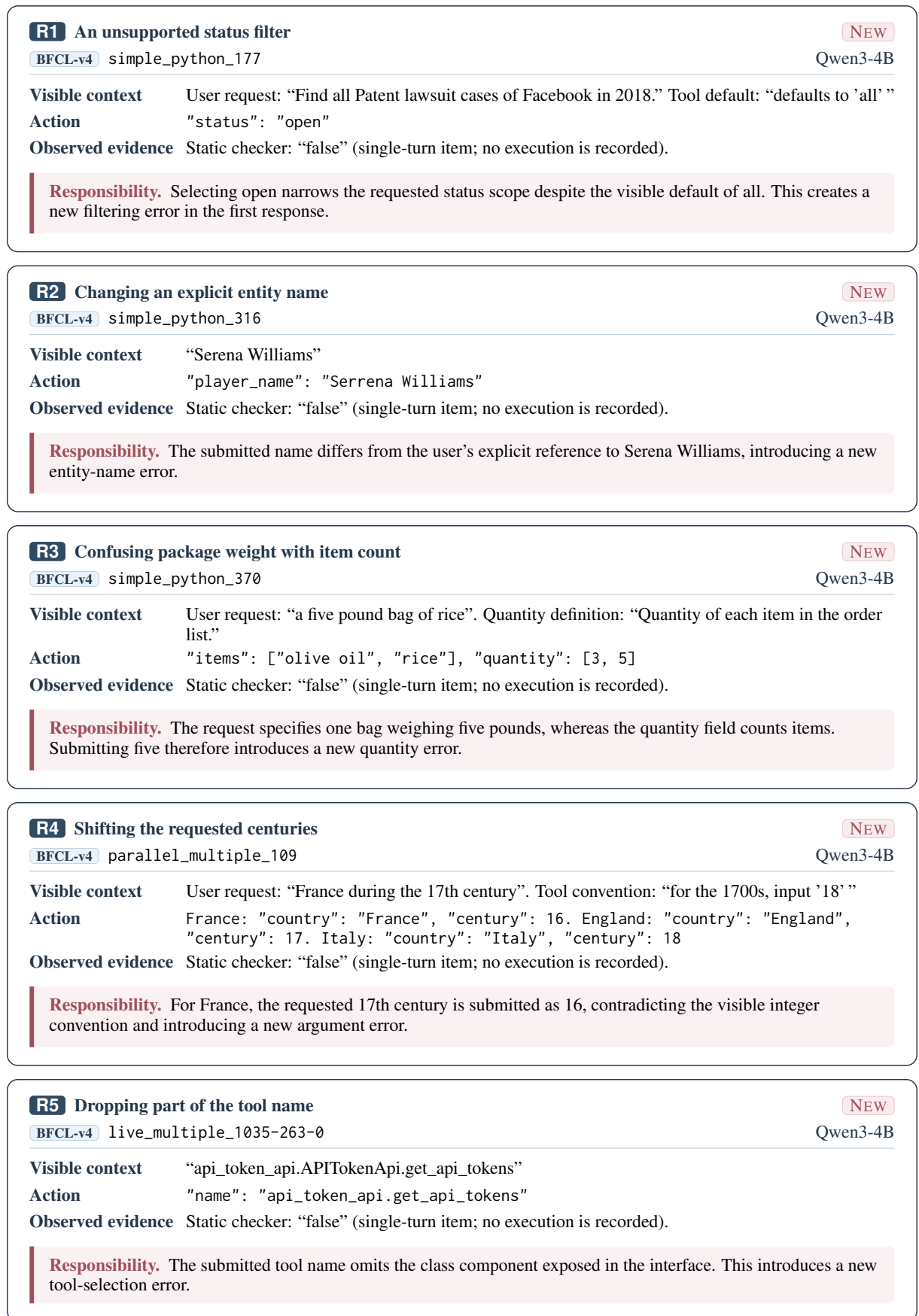

\centering
\ResponsibilityCase{R1}{An unsupported status filter}
{\BaseMeta{BFCL-v4}{simple\_\allowbreak{}python\_\allowbreak{}177}{Qwen3-4B}}
{User request: ``Find all Patent lawsuit cases of Facebook in 2018.'' Tool default: ``defaults to 'all'\,'' }
{"status": "open"}
{Static checker: ``false'' (single-turn item; no execution is recorded).}
{Selecting open narrows the requested status scope despite the visible default of all. This creates a new filtering error in the first response.}{New}

\ResponsibilityCase{R2}{Changing an explicit entity name}
{\BaseMeta{BFCL-v4}{simple\_\allowbreak{}python\_\allowbreak{}316}{Qwen3-4B}}
{``Serena Williams'' }
{"player\_\allowbreak{}name": "Serrena Williams"}
{Static checker: ``false'' (single-turn item; no execution is recorded).}
{The submitted name differs from the user's explicit reference to Serena Williams, introducing a new entity-name error.}{New}

\ResponsibilityCase{R3}{Confusing package weight with item count}
{\BaseMeta{BFCL-v4}{simple\_\allowbreak{}python\_\allowbreak{}370}{Qwen3-4B}}
{User request: ``a five pound bag of rice''. Quantity definition: ``Quantity of each item in the order list.'' }
{"items": ["olive oil", "rice"], "quantity": [3, 5]}
{Static checker: ``false'' (single-turn item; no execution is recorded).}
{The request specifies one bag weighing five pounds, whereas the quantity field counts items. Submitting five therefore introduces a new quantity error.}{New}

\ResponsibilityCase{R4}{Shifting the requested centuries}
{\BaseMeta{BFCL-v4}{parallel\_\allowbreak{}multiple\_\allowbreak{}109}{Qwen3-4B}}
{User request: ``France during the 17th century''. Tool convention: ``for the 1700s, input '18'\,'' }
{France: "country": "France", "century": 16. England: "country": "England", "century": 17. Italy: "country": "Italy", "century": 18}
{Static checker: ``false'' (single-turn item; no execution is recorded).}
{For France, the requested 17th century is submitted as 16, contradicting the visible integer convention and introducing a new argument error.}{New}

\ResponsibilityCase{R5}{Dropping part of the tool name}
{\BaseMeta{BFCL-v4}{live\_\allowbreak{}multiple\_\allowbreak{}1035-263-0}{Qwen3-4B}}
{``api\_\allowbreak{}token\_\allowbreak{}api.APITokenApi.get\_\allowbreak{}api\_\allowbreak{}tokens'' }
{"name": "api\_\allowbreak{}token\_\allowbreak{}api.get\_\allowbreak{}api\_\allowbreak{}tokens"}
{Static checker: ``false'' (single-turn item; no execution is recorded).}
{The submitted tool name omits the class component exposed in the interface. This introduces a new tool-selection error.}{New}
\vspace{-5pt}
\caption{Base-model \textsc{New} events.}
\label{fig:case-base-new}
\end{figure*}

\begin{figure*}[p]
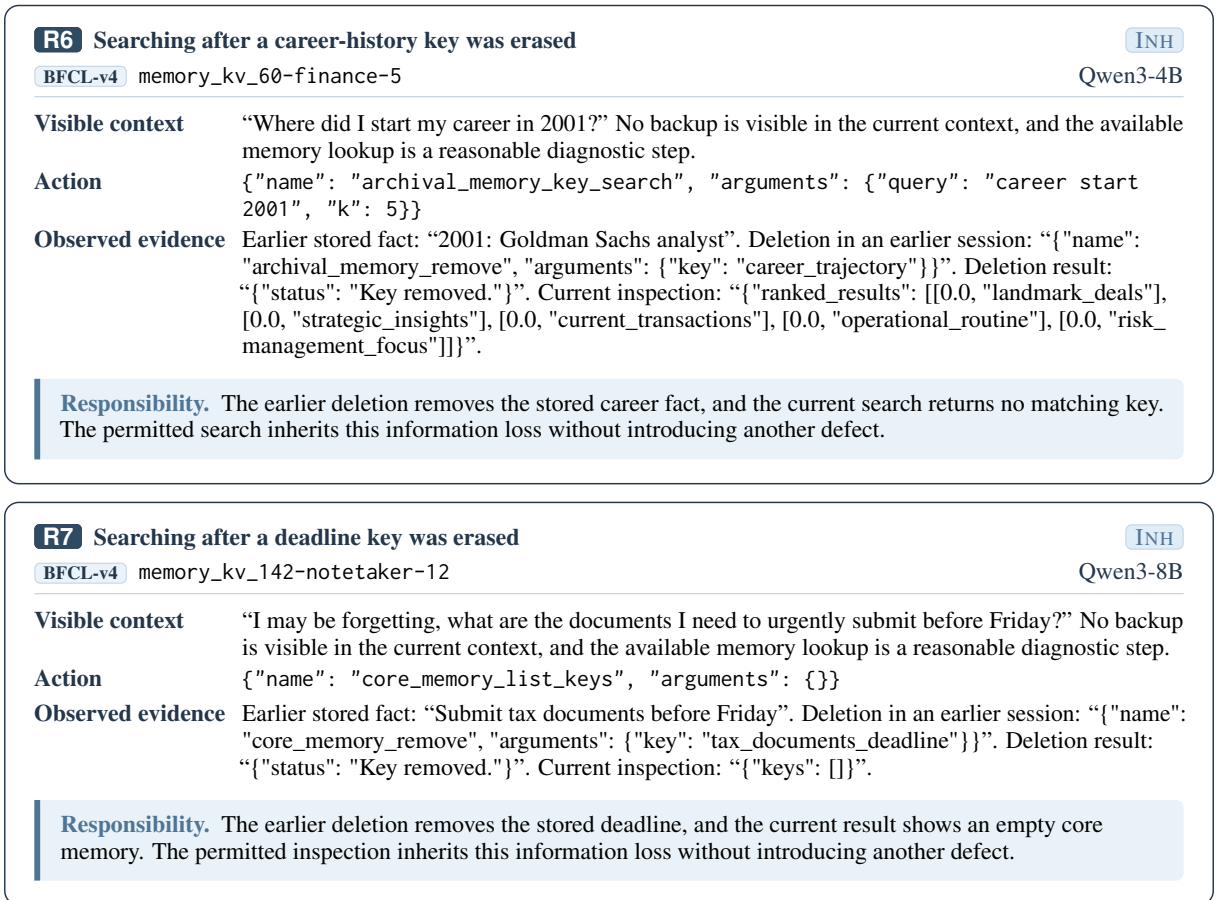

\centering
\ResponsibilityCase{R6}{Searching after a career-history key was erased}
{\BaseMeta{BFCL-v4}{memory\_\allowbreak{}kv\_\allowbreak{}60-finance-5}{Qwen3-4B}}
{``Where did I start my career in 2001?'' No backup is visible in the current context, and the available memory lookup is a reasonable diagnostic step.}
{\{"name": "archival\_\allowbreak{}memory\_\allowbreak{}key\_\allowbreak{}search", "arguments": \{"query": "career start 2001", "k": 5\}\}}
{Earlier stored fact: ``2001: Goldman Sachs analyst''. Deletion in an earlier session: ``\{"name": "archival\_\allowbreak{}memory\_\allowbreak{}remove", "arguments": \{"key": "career\_\allowbreak{}trajectory"\}\}''. Deletion result: ``\{"status": "Key removed."\}''. Current inspection: ``\{"ranked\_\allowbreak{}results": [[0.0, "landmark\_\allowbreak{}deals"], [0.0, "strategic\_\allowbreak{}insights"], [0.0, "current\_\allowbreak{}transactions"], [0.0, "operational\_\allowbreak{}routine"], [0.0, "risk\_\allowbreak{}management\_\allowbreak{}focus"]]\}''.}
{The earlier deletion removes the stored career fact, and the current search returns no matching key. The permitted search inherits this information loss without introducing another defect.}{Inh}

\ResponsibilityCase{R7}{Searching after a deadline key was erased}
{\BaseMeta{BFCL-v4}{memory\_\allowbreak{}kv\_\allowbreak{}142-notetaker-12}{Qwen3-8B}}
{``I may be forgetting, what are the documents I need to urgently submit before Friday?'' No backup is visible in the current context, and the available memory lookup is a reasonable diagnostic step.}
{\{"name": "core\_\allowbreak{}memory\_\allowbreak{}list\_\allowbreak{}keys", "arguments": \{\}\}}
{Earlier stored fact: ``Submit tax documents before Friday''. Deletion in an earlier session: ``\{"name": "core\_\allowbreak{}memory\_\allowbreak{}remove", "arguments": \{"key": "tax\_\allowbreak{}documents\_\allowbreak{}deadline"\}\}''. Deletion result: ``\{"status": "Key removed."\}''. Current inspection: ``\{"keys": []\}''.}
{The earlier deletion removes the stored deadline, and the current result shows an empty core memory. The permitted inspection inherits this information loss without introducing another defect.}{Inh}
\vspace{-5pt}
\caption{Base-model \textsc{Inh} events.}
\label{fig:case-base-inh}
\end{figure*}

\begin{figure*}[p]
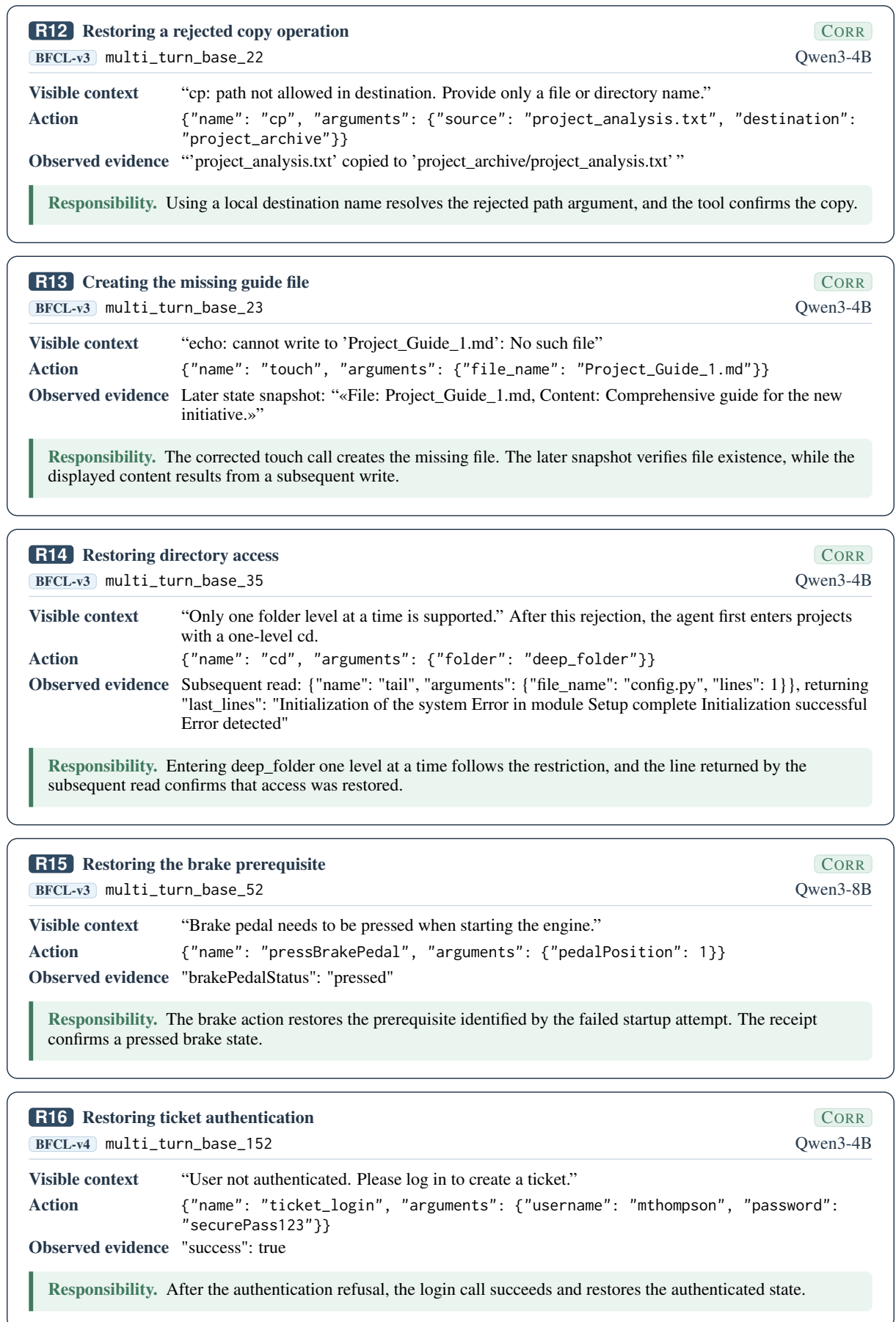

\centering
\ResponsibilityCase{R12}{Restoring a rejected copy operation}
{\BaseMeta{BFCL-v3}{multi\_\allowbreak{}turn\_\allowbreak{}base\_\allowbreak{}22}{Qwen3-4B}}
{``cp: path not allowed in destination. Provide only a file or directory name.'' }
{\{"name": "cp", "arguments": \{"source": "project\_\allowbreak{}analysis.txt", "destination": "project\_\allowbreak{}archive"\}\}}
{``'project\_\allowbreak{}analysis.txt' copied to 'project\_\allowbreak{}archive/project\_\allowbreak{}analysis.txt'\,'' }
{Using a local destination name resolves the rejected path argument, and the tool confirms the copy.}{Corr}

\ResponsibilityCase{R13}{Creating the missing guide file}
{\BaseMeta{BFCL-v3}{multi\_\allowbreak{}turn\_\allowbreak{}base\_\allowbreak{}23}{Qwen3-4B}}
{``echo: cannot write to 'Project\_\allowbreak{}Guide\_\allowbreak{}1.md': No such file'' }
{\{"name": "touch", "arguments": \{"file\_\allowbreak{}name": "Project\_\allowbreak{}Guide\_\allowbreak{}1.md"\}\}}
{Later state snapshot: ``<<File: Project\_\allowbreak{}Guide\_\allowbreak{}1.md, Content: Comprehensive guide for the new initiative.>>''}
{The corrected touch call creates the missing file. The later snapshot verifies file existence, while the displayed content results from a subsequent write.}{Corr}

\ResponsibilityCase{R14}{Restoring directory access}
{\BaseMeta{BFCL-v3}{multi\_\allowbreak{}turn\_\allowbreak{}base\_\allowbreak{}35}{Qwen3-4B}}
{``Only one folder level at a time is supported.'' After this rejection, the agent first enters projects with a one-level cd.}
{\{"name": "cd", "arguments": \{"folder": "deep\_\allowbreak{}folder"\}\}}
{Subsequent read: \{"name": "tail", "arguments": \{"file\_\allowbreak{}name": "config.py", "lines": 1\}\}\textrm{, returning }"last\_\allowbreak{}lines": "Initialization of the system Error in module Setup complete Initialization successful Error detected" }
{Entering deep\_\allowbreak{}folder one level at a time follows the restriction, and the line returned by the subsequent read confirms that access was restored.}{Corr}

\ResponsibilityCase{R15}{Restoring the brake prerequisite}
{\BaseMeta{BFCL-v3}{multi\_\allowbreak{}turn\_\allowbreak{}base\_\allowbreak{}52}{Qwen3-8B}}
{``Brake pedal needs to be pressed when starting the engine.'' }
{\{"name": "pressBrakePedal", "arguments": \{"pedalPosition": 1\}\}}
{"brakePedalStatus": "pressed" }
{The brake action restores the prerequisite identified by the failed startup attempt. The receipt confirms a pressed brake state.}{Corr}

\ResponsibilityCase{R16}{Restoring ticket authentication}
{\BaseMeta{BFCL-v4}{multi\_\allowbreak{}turn\_\allowbreak{}base\_\allowbreak{}152}{Qwen3-4B}}
{``User not authenticated. Please log in to create a ticket.'' }
{\{"name": "ticket\_\allowbreak{}login", "arguments": \{"username": "mthompson", "password": "securePass123"\}\}}
{"success": true }
{After the authentication refusal, the login call succeeds and restores the authenticated state.}{Corr}
\vspace{-5pt}
\caption{Base-model \textsc{Corr} events.}
\label{fig:case-base-corr}
\end{figure*}

Figures~\ref{fig:case-base-new}--\ref{fig:case-base-corr} illustrate each event type in Qwen3 base-model trajectories.
In these panels, \textsc{New} events arise when a submitted call contradicts a visible user constraint or tool contract.
\textsc{Inh} events arise when the agent searches or inspects memory for a fact deleted in an earlier session, with neither the fact nor its deletion visible in its context.
\textsc{Miss} events arise when the agent repeats a rejected call unchanged although the exposed discrepancy admits a feasible response.
\textsc{Corr} events arise when a tool result or state snapshot confirms that a corrective call restored a violated requirement or precondition.
Because events are assessed per action and predicate, one trajectory can contain a \textsc{Miss} event on one requirement and a \textsc{Corr} event on another.

\subsection{Paired Comparisons Before and After ECHO Training}
\label{app:paired-cases}

\begin{figure*}[p]
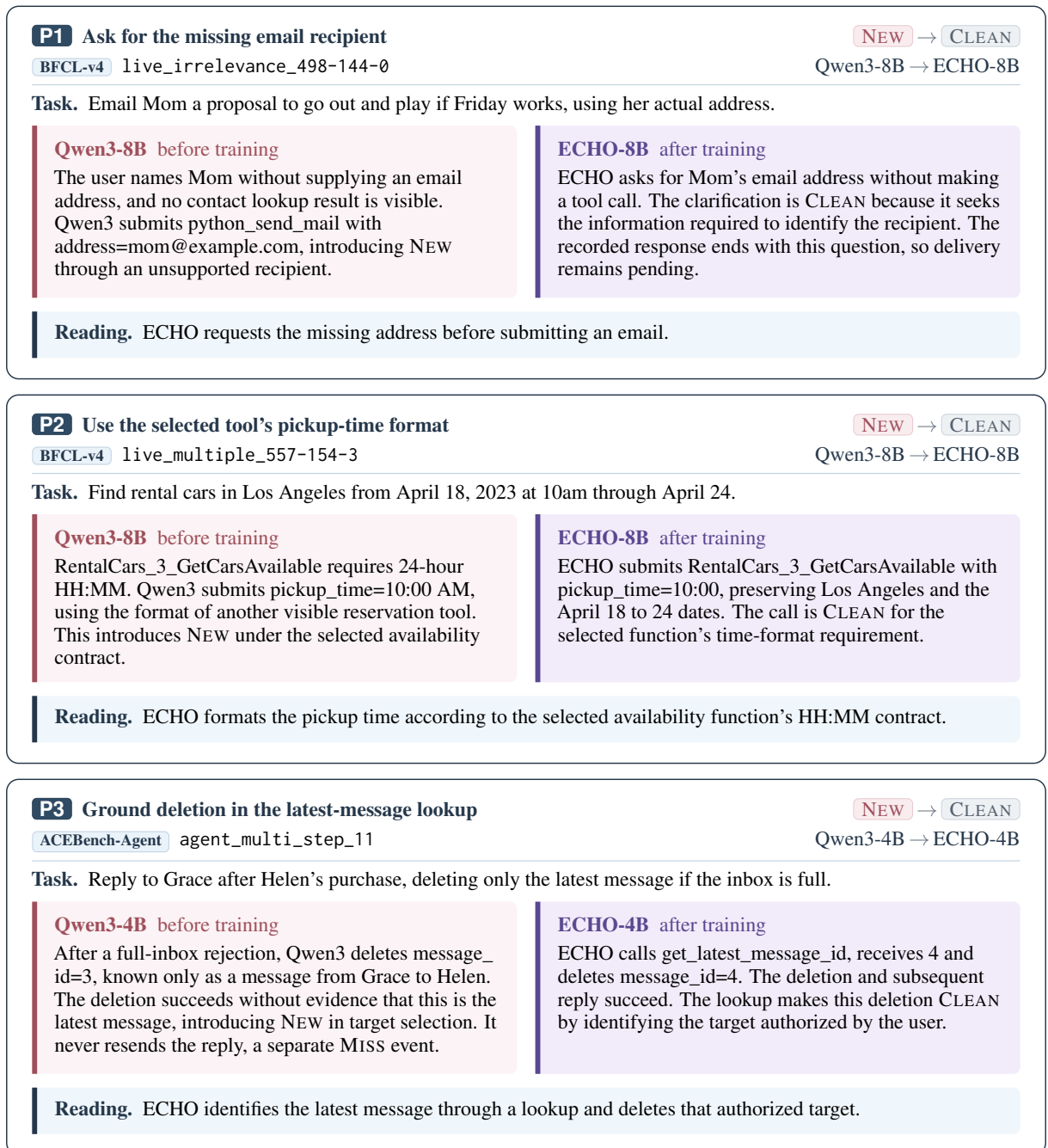

\centering
\ComparisonCase{P1}{Ask for the missing email recipient}
{\PairMeta{BFCL-v4}{live\_\allowbreak{}irrelevance\_\allowbreak{}498-144-0}{8B}}
{Email Mom a proposal to go out and play if Friday works, using her actual address.}
{The user names Mom without supplying an email address, and no contact lookup result is visible. Qwen3 submits python\_\allowbreak{}send\_\allowbreak{}mail with address=mom@example.com, introducing \textsc{New} through an unsupported recipient.}
{ECHO asks for Mom's email address without making a tool call. The clarification is \textsc{Clean} because it seeks the information required to identify the recipient. The recorded response ends with this question, so delivery remains pending.}
{ECHO requests the missing address before submitting an email.}{New}{Clean}

\ComparisonCase{P2}{Use the selected tool's pickup-time format}
{\PairMeta{BFCL-v4}{live\_\allowbreak{}multiple\_\allowbreak{}557-154-3}{8B}}
{Find rental cars in Los Angeles from April 18, 2023 at 10am through April 24.}
{RentalCars\_\allowbreak{}3\_\allowbreak{}GetCarsAvailable requires 24-hour HH:MM. Qwen3 submits pickup\_\allowbreak{}time=10:00 AM, using the format of another visible reservation tool. This introduces \textsc{New} under the selected availability contract.}
{ECHO submits RentalCars\_\allowbreak{}3\_\allowbreak{}GetCarsAvailable with pickup\_\allowbreak{}time=10:00, preserving Los Angeles and the April 18 to 24 dates. The call is \textsc{Clean} for the selected function's time-format requirement.}
{ECHO formats the pickup time according to the selected availability function's HH:MM contract.}{New}{Clean}

\ComparisonCase{P3}{Ground deletion in the latest-message lookup}
{\PairMeta{ACEBench-Agent}{agent\_\allowbreak{}multi\_\allowbreak{}step\_\allowbreak{}11}{4B}}
{Reply to Grace after Helen's purchase, deleting only the latest message if the inbox is full.}
{After a full-inbox rejection, Qwen3 deletes message\_\allowbreak{}id=3, known only as a message from Grace to Helen. The deletion succeeds without evidence that this is the latest message, introducing \textsc{New} in target selection. It never resends the reply, a separate \textsc{Miss} event.}
{ECHO calls get\_\allowbreak{}latest\_\allowbreak{}message\_\allowbreak{}id, receives 4 and deletes message\_\allowbreak{}id=4. The deletion and subsequent reply succeed. The lookup makes this deletion \textsc{Clean} by identifying the target authorized by the user.}
{ECHO identifies the latest message through a lookup and deletes that authorized target.}{New}{Clean}
\vspace{-5pt}
\caption{Paired comparisons of Qwen3 (before training) and ECHO (after training) on the same tasks.}
\label{fig:case-pair-1}
\end{figure*}

\begin{figure*}[p]
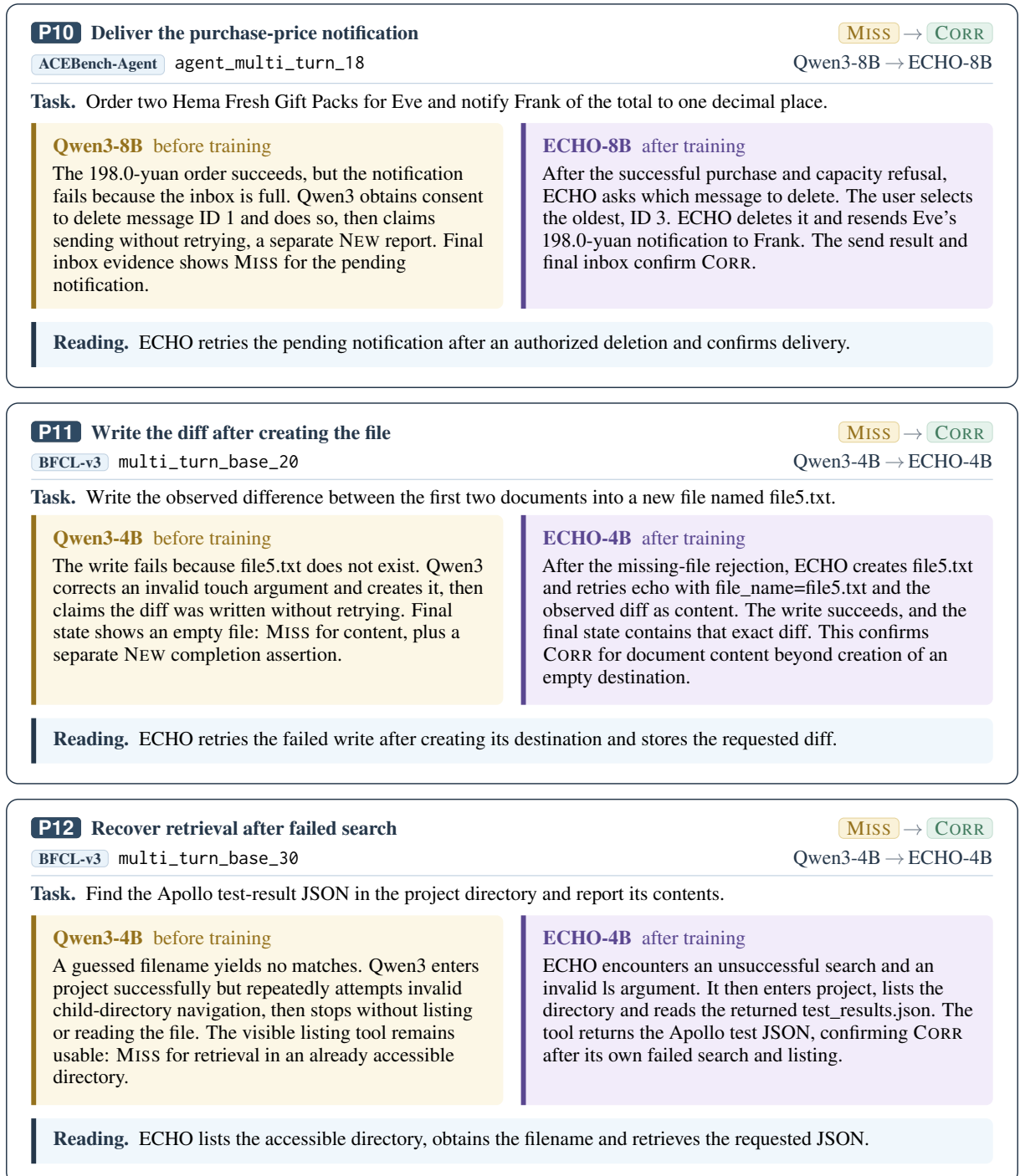

\centering
\ComparisonCase{P10}{Deliver the purchase-price notification}
{\PairMeta{ACEBench-Agent}{agent\_\allowbreak{}multi\_\allowbreak{}turn\_\allowbreak{}18}{8B}}
{Order two Hema Fresh Gift Packs for Eve and notify Frank of the total to one decimal place.}
{The 198.0-yuan order succeeds, but the notification fails because the inbox is full. Qwen3 obtains consent to delete message ID 1 and does so, then claims sending without retrying, a separate \textsc{New} report. Final inbox evidence shows \textsc{Miss} for the pending notification.}
{After the successful purchase and capacity refusal, ECHO asks which message to delete. The user selects the oldest, ID 3. ECHO deletes it and resends Eve's 198.0-yuan notification to Frank. The send result and final inbox confirm \textsc{Corr}.}
{ECHO retries the pending notification after an authorized deletion and confirms delivery.}{Miss}{Corr}

\ComparisonCase{P11}{Write the diff after creating the file}
{\PairMeta{BFCL-v3}{multi\_\allowbreak{}turn\_\allowbreak{}base\_\allowbreak{}20}{4B}}
{Write the observed difference between the first two documents into a new file named file5.txt.}
{The write fails because file5.txt does not exist. Qwen3 corrects an invalid touch argument and creates it, then claims the diff was written without retrying. Final state shows an empty file: \textsc{Miss} for content, plus a separate \textsc{New} completion assertion.}
{After the missing-file rejection, ECHO creates file5.txt and retries echo with file\_\allowbreak{}name=file5.txt and the observed diff as content. The write succeeds, and the final state contains that exact diff. This confirms \textsc{Corr} for document content beyond creation of an empty destination.}
{ECHO retries the failed write after creating its destination and stores the requested diff.}{Miss}{Corr}

\ComparisonCase{P12}{Recover retrieval after failed search}
{\PairMeta{BFCL-v3}{multi\_\allowbreak{}turn\_\allowbreak{}base\_\allowbreak{}30}{4B}}
{Find the Apollo test-result JSON in the project directory and report its contents.}
{A guessed filename yields no matches. Qwen3 enters project successfully but repeatedly attempts invalid child-directory navigation, then stops without listing or reading the file. The visible listing tool remains usable: \textsc{Miss} for retrieval in an already accessible directory.}
{ECHO encounters an unsuccessful search and an invalid ls argument. It then enters project, lists the directory and reads the returned test\_\allowbreak{}results.json. The tool returns the Apollo test JSON, confirming \textsc{Corr} after its own failed search and listing.}
{ECHO lists the accessible directory, obtains the filename and retrieves the requested JSON.}{Miss}{Corr}
\vspace{-5pt}
\caption{Paired comparisons of Qwen3 and ECHO on the same tasks (continued).}
\label{fig:case-pair-4}
\end{figure*}

Figures~\ref{fig:case-pair-1}--\ref{fig:case-pair-5} compare Qwen3 and ECHO at the same scale on the same task, complementing \S\ref{sec:case-study}.
Paired runs share the evaluation harness, tool contracts, decoding configuration, environment settings, and sampling seed.
Because interaction histories can diverge, each comparison relates both trajectories to the same requirement rather than to aligned action indices.
In pairs where Qwen3 introduces a \textsc{New} error, ECHO's action on the same requirement is \textsc{Clean}.
Where Qwen3's downstream response inherits information lost through earlier deletion or omitted persistence, ECHO preserves the required fact upstream, even though \textsc{Inh} events receive no local correction (\S\ref{sec:budgets}).
Where Qwen3 misses a correction, sometimes after completing a prerequisite repair such as logging in or freeing inbox capacity, ECHO either avoids the discrepancy or completes the pending operation, with restoration verified by tool results or the final state.

\renewcommand{\dblfloatpagefraction}{0.01}
\makeatletter\setlength{\@dblfpsep}{2\textheight}\setlength{\@dblfptop}{0pt plus 1fil}\setlength{\@dblfpbot}{0pt plus 1fil}\makeatother
\begin{figure*}[p]
\centering
\ResponsibilityCase{R8}{Retrying cruise activation with the engine stopped}
{\BaseMeta{BFCL-v3}{multi\_\allowbreak{}turn\_\allowbreak{}base\_\allowbreak{}79}{Qwen3-4B}}
{``\{"error": "Start the engine before activating the cruise control."\}'' A feasible response was to address the engine prerequisites or explain the blockage and pause cruise activation.}
{\{"name": "setCruiseControl", "arguments": \{"speed": 65, "activate": true, "distanceToNextVehicle": 100\}\}}
{``\{"error": "Start the engine before activating the cruise control."\}'' }
{The repeated cruise call leaves the exposed engine prerequisite unaddressed despite an available response.}{Miss}

\ResponsibilityCase{R9}{Repeating a write to a missing file}
{\BaseMeta{BFCL-v3}{multi\_\allowbreak{}turn\_\allowbreak{}base\_\allowbreak{}35}{Qwen3-4B}}
{``\{"error": "echo: cannot write to 'diff.txt': No such file"\}'' The available touch tool permits creating diff.txt before retrying the write.}
{Function: "name": "echo". Target argument: "file\_\allowbreak{}name": "diff.txt"}
{``\{"error": "echo: cannot write to 'diff.txt': No such file"\}'' }
{The agent repeats the rejected write without first creating the missing file, leaving an actionable prerequisite unresolved.}{Miss}

\ResponsibilityCase{R10}{Retrying an undefined market-status function}
{\BaseMeta{BFCL-v3}{multi\_\allowbreak{}turn\_\allowbreak{}base\_\allowbreak{}112}{Qwen3-4B}}
{``Error during execution: name 'update\_\allowbreak{}market\_\allowbreak{}status' is not defined'' The agent could explain that the requested operation is unavailable and stop the unchanged retries.}
{\{"name": "update\_\allowbreak{}market\_\allowbreak{}status", "arguments": \{"current\_\allowbreak{}time\_\allowbreak{}str": "10:30 AM"\}\}}
{``Error during execution: name 'update\_\allowbreak{}market\_\allowbreak{}status' is not defined'' }
{The unchanged retry follows an explicit undefined-function error and leaves the execution limitation unaddressed.}{Miss}

\ResponsibilityCase{R11}{Repeating an undefined withdrawal operation}
{\BaseMeta{BFCL-v3}{multi\_\allowbreak{}turn\_\allowbreak{}base\_\allowbreak{}121}{Qwen3-4B}}
{``Error during execution: name 'make\_\allowbreak{}transaction' is not defined'' The agent could report the execution failure and pause the withdrawal instead of repeating the same call.}
{\{"name": "make\_\allowbreak{}transaction", "arguments": \{"account\_\allowbreak{}id": 12345, "xact\_\allowbreak{}type": "withdrawal", "amount": 500\}\}}
{``Error during execution: name 'make\_\allowbreak{}transaction' is not defined'' }
{The agent repeats the unavailable withdrawal operation despite the visible failure and a feasible response to the blockage.}{Miss}
\vspace{-5pt}
\caption{Base-model \textsc{Miss} events.}
\label{fig:case-base-miss}
\end{figure*}
\begin{figure*}[p]
\centering
\ComparisonCase{P4}{Preserve the investment firm identity}
{\PairMeta{BFCL-v4}{memory\_\allowbreak{}rec\_\allowbreak{}sum\_\allowbreak{}55-finance-0}{4B}}
{Retain the user profile across sessions and later identify the investment firm the user manages.}
{An earlier memory\_\allowbreak{}append stores Legend Investments, but a later memory\_\allowbreak{}clear erases it. The firm name is absent from the answer prefix, so Qwen3 reasonably declines to supply it. The downstream information loss is \textsc{Inh} from that clearing operation, without a current factual \textsc{New} or \textsc{Miss}.}
{ECHO's earlier memory\_\allowbreak{}append successfully stores Legend Investments. The name remains available in downstream memory, allowing ECHO to answer Legend Investments. This answer is \textsc{Clean} because the fact was retained, with no repair required at the answer step.}
{Preserving the upstream firm-name record makes the requested investment identity available for the later answer.}{Inh}{Clean}

\ComparisonCase{P5}{Preserve the morning start time}
{\PairMeta{BFCL-v4}{memory\_\allowbreak{}rec\_\allowbreak{}sum\_\allowbreak{}64-finance-9}{8B}}
{Retain the daily schedule and later report the morning start time in HH:MM AM/PM format.}
{An earlier memory\_\allowbreak{}update stores the 5:30 AM schedule, but a later memory\_\allowbreak{}clear erases it. Qwen3 reasonably declines to provide the time because it is absent from the answer prefix. The missing information is \textsc{Inh} from the recorded clearing operation.}
{ECHO's earlier memory\_\allowbreak{}append successfully stores the 5:30 AM schedule. The fact remains in downstream memory, and ECHO answers 05:30 AM in the requested format. This \textsc{Clean} answer uses retained information without repairing memory at the target step.}
{The retained schedule supports the requested morning time and its HH:MM AM/PM representation.}{Inh}{Clean}

\ComparisonCase{P6}{Persist the teacher's counting range}
{\PairMeta{BFCL-v4}{memory\_\allowbreak{}rec\_\allowbreak{}sum\_\allowbreak{}154-notetaker-24}{4B}}
{Remember the family notes and later report the teacher's recommended counting range as X-Y.}
{The user supplies counting numbers 1-20. Qwen3 claims to have added the family plan but executes no memory write, a separate \textsc{New} claim. Later memory lacks the range, and it reasonably declines to supply it. This \textsc{Inh} consequence originates in omitted persistence.}
{ECHO executes memory\_\allowbreak{}append with the family plan and counting range 1-20, and the tool confirms persistence. The range remains available in subsequent memory, allowing ECHO to answer 1-20 from its prefix. The answer is \textsc{Clean} through successful storage rather than repair at the target step.}
{ECHO's confirmed memory append preserves the requested range for subsequent recall.}{Inh}{Clean}
\vspace{-5pt}
\caption{Paired comparisons of Qwen3 and ECHO on the same tasks (continued).}
\label{fig:case-pair-2}
\end{figure*}
\begin{figure*}[p]
\centering
\ComparisonCase{P7}{Prepare the archive before searching}
{\PairMeta{BFCL-v4}{multi\_\allowbreak{}turn\_\allowbreak{}base\_\allowbreak{}36}{4B}}
{Create project\_\allowbreak{}summary.txt, copy it into archive as summary\_\allowbreak{}2024.txt, and report lines containing Progress.}
{After missing-file and navigation errors, Qwen3 repeats cd, cp, cd and grep without resolving its location: \textsc{Miss} despite available pwd or ls. Later attempts fail, no archived summary appears, and its final answer invents matching lines, a separate \textsc{New} report.}
{ECHO creates the file, copies it into archive, enters that directory and renames the copy summary\_\allowbreak{}2024.txt. The grep search for Progress returns no matches, which ECHO reports accurately. Both empty files remain in the final state. The search is \textsc{Clean} because the required file was prepared before retrieval.}
{ECHO prepares the requested archive file and reports the observed search result.}{Miss}{Clean}

\ComparisonCase{P8}{Place the order after login}
{\PairMeta{ACEBench-Agent}{agent\_\allowbreak{}multi\_\allowbreak{}step\_\allowbreak{}3}{8B}}
{Check Jiutian BBQ's menu and place one order Jack can afford within 120 yuan.}
{The menu offers Grilled Pork Belly for 78 yuan, but ordering fails because Jack is logged out. After successful login, Qwen3 declares completion without retrying. Final orders remain empty: \textsc{Miss} for the pending purchase, with a separate \textsc{New} completion claim.}
{After the rejected order and successful login, ECHO retries add\_\allowbreak{}food\_\allowbreak{}delivery\_\allowbreak{}order for Jack at Jiutian BBQ with one Grilled Pork Belly. The receipt confirms 78 yuan, and the final state contains one order and a 42-yuan balance. These observations verify \textsc{Corr}.}
{ECHO retries the affordable order after authentication and obtains a purchase receipt.}{Miss}{Corr}

\ComparisonCase{P9}{Send the reply after freeing capacity}
{\PairMeta{ACEBench-Agent}{agent\_\allowbreak{}multi\_\allowbreak{}step\_\allowbreak{}12}{4B}}
{Buy Grace's requested cheaper Heytea milk tea and reply, using the user's authorization to delete the latest message if needed.}
{Qwen3 buys the milk tea, but its reply is rejected because the inbox is full. It retrieves latest message ID 4 and deletes it, then stops without resending. The final inbox lacks the confirmation: \textsc{Miss} for the pending reply.}
{ECHO also buys the milk tea and encounters the reply rejection. After deleting the authorized latest message, it retries send\_\allowbreak{}message from Helen to Grace with message=Already bought. The tool result and final inbox confirm that exact reply: \textsc{Corr}.}
{ECHO resends the rejected purchase reply after the authorized deletion.}{Miss}{Corr}
\vspace{-5pt}
\caption{Paired comparisons of Qwen3 and ECHO on the same tasks (continued).}
\label{fig:case-pair-3}
\end{figure*}
\begin{figure*}[p]
\centering
\ComparisonCase{P13}{Retry ignition after prerequisite repair}
{\PairMeta{BFCL-v3}{multi\_\allowbreak{}turn\_\allowbreak{}base\_\allowbreak{}56}{4B}}
{Start the engine and ensure the vehicle doors are locked after the requested refuelling.}
{Qwen3 issues START before locking the doors. Ignition is rejected, then door locking succeeds. It stops without an authorized START retry, leaving the final engine stopped. This is \textsc{Miss} for ignition after repairing the visible door prerequisite.}
{After ignition is rejected because the doors are unlocked, ECHO locks the doors and retries START. A further refusal identifies the brake prerequisite, prompting ECHO to fully press the brake and retry ignition. The tool result and final state confirm a running engine, establishing \textsc{Corr}.}
{ECHO resolves each reported ignition prerequisite and verifies engine startup through execution.}{Miss}{Corr}

\ComparisonCase{P14}{Repair timestamp bounds to retrieve the oldest message}
{\PairMeta{ToolSandbox}{search\_\allowbreak{}message\_\allowbreak{}with\_\allowbreak{}recency\_\allowbreak{}oldest\_\allowbreak{}all\_\allowbreak{}tools}{4B}}
{Retrieve the oldest recorded message and report the content associated with its timestamp.}
{After an empty search is rejected, the user says identifying details are unavailable. Qwen3 repeatedly asks for those details and performs no further search: \textsc{Miss} despite visible timestamp filtering and a current-time tool. The initial clarification was reasonable.}
{ECHO's empty search and two timestamp ranges are rejected. It uses the reported bounds to retry with creation\_\allowbreak{}timestamp\_\allowbreak{}lowerbound=315529200 and creation\_\allowbreak{}timestamp\_\allowbreak{}upperbound=2524604400. The linked result returns five messages, from which ECHO reports the content at the smallest timestamp. This confirms \textsc{Corr}.}
{ECHO repairs the timestamp bounds using tool feedback and retrieves the oldest message.}{Miss}{Corr}

\ComparisonCase{P15}{Retry the return with the authorized card}
{\PairMeta{$\tau^2$-Bench}{retail\_\allowbreak{}13}{4B}}
{Register the three-item return for order \#W5490111, keeping the keyboard and mouse.}
{The PayPal return is rejected. After the user explicitly approves the original credit card, Qwen3 submits an empty return on another order and claims that the target return succeeded, a separate \textsc{New} report. It never retries \#W5490111. \textsc{Miss} applies after authorization, while the earlier confirmation question was reasonable.}
{After the refund-method rejection, ECHO obtains explicit approval and retries \#W5490111 with the same three items and payment\_\allowbreak{}method\_\allowbreak{}id=credit\_\allowbreak{}card\_\allowbreak{}3124723. The linked result confirms registration of the return request with that item set and card, establishing \textsc{Corr}. No later tool call reverses the registration.}
{ECHO obtains approval for the permitted card, retries the original order and items, and confirms registration of the return request.}{Miss}{Corr}
\vspace{-5pt}
\caption{Paired comparisons of Qwen3 and ECHO on the same tasks (continued).}
\label{fig:case-pair-5}
\end{figure*}

\end{document}